\pdfoutput=1

\documentclass[11pt]{article}
\usepackage[dvipsnames]{xcolor}
\usepackage[final]{acl}
\usepackage{times}
\usepackage{alltt}
\usepackage{latexsym}
\usepackage{amsmath,amssymb,bm} 
\DeclareMathOperator*{\argmax}{arg\,max}
\usepackage[textsize=scriptsize,disable]{todonotes}
\usepackage[inline,shortlabels]{enumitem} \setlist{nosep}

\usepackage[T1]{fontenc}

\usepackage[utf8]{inputenc}
\usepackage[russian,english]{babel}  
\usepackage{microtype}

\usepackage{inconsolata}

\usepackage{graphicx}
\usepackage{courier}  
\usepackage{multirow}
\usepackage{booktabs}
\usepackage{color}  
\usepackage{verbatim}  
\usepackage{adjustbox} 
\usepackage{listings}
\definecolor{delim}{RGB}{20,105,176}
\definecolor{numb}{RGB}{106, 109, 32}
\definecolor{string}{rgb}{0.64,0.08,0.08}

\lstdefinelanguage{json}{
    numbers=left,
    numberstyle=\small,
    frame=single,
    rulecolor=\color{black},
    showspaces=false,
    showtabs=false,
    breaklines=true,
    postbreak=\raisebox{0ex}[0ex][0ex]{\ensuremath{\color{gray}\hookrightarrow\space}},
    breakatwhitespace=true,
    basicstyle=\ttfamily\small,
    upquote=true,
    morestring=[b]",
    stringstyle=\color{string},
    literate=
     *{0}{{{\color{numb}0}}}{1}
      {1}{{{\color{numb}1}}}{1}
      {2}{{{\color{numb}2}}}{1}
      {3}{{{\color{numb}3}}}{1}
      {4}{{{\color{numb}4}}}{1}
      {5}{{{\color{numb}5}}}{1}
      {6}{{{\color{numb}6}}}{1}
      {7}{{{\color{numb}7}}}{1}
      {8}{{{\color{numb}8}}}{1}
      {9}{{{\color{numb}9}}}{1}
      {\{}{{{\color{delim}{\{}}}}{1}
      {\}}{{{\color{delim}{\}}}}}{1}
      {[}{{{\color{delim}{[}}}}{1}
      {]}{{{\color{delim}{]}}}}{1},
}

\usepackage{framed}
\usepackage{anyfontsize}
\usepackage{tcolorbox}
\tcbuselibrary{skins}
\tcbuselibrary{breakable}
\usepackage[skip=.1ex plus.1ex]{caption}
\usepackage{listings}

\usepackage[noEnd=true,indLines=true,commentColor=Periwinkle,beginComment=/*~, endComment=~*/,beginLComment=/*~, endLComment=~*/]{algpseudocodex}

\usepackage{colortbl}
\usepackage{soul}
\usepackage[utf8]{inputenc}
\usepackage[T1]{fontenc}
\usepackage{listingsutf8}
\usepackage{capt-of}

\usepackage{algorithm}

\makeatletter
\let\ftype@table\ftype@figure
\g@addto@macro{\normalsize}{%
\setlength{\abovedisplayskip}{2pt minus1pt}%
\setlength{\abovedisplayshortskip}{2pt minus1pt}%
\setlength{\belowdisplayskip}{2pt minus1pt}%
\setlength{\belowdisplayshortskip}{2pt minus1pt}}

\textfloatsep \intextsep
\floatsep \intextsep
\makeatother

\newcommand{\AugD}{\mathcal{T}_\text{D}}
\newcommand{\train}{\mathcal{T}}

\newcommand{\sysname}{SCUD4ICL}
\def\longname{sub code+utterance decomposition for in-context learning}
\newcommand{\coverLS}{CoverLS}
\newcommand{\baseline}{\coverLS}
\newcommand{\indp}{Independent}
\newcommand{\vx}{\mathbf{x}}
\newcommand{\vy}{\mathbf{y}}

\newcommand{\chatgpt}{GPT3.5}
\newcommand{\gpt}{GPT4o}
\newcommand{\mistral}{Mistral:7b}
\newcommand{\llama}{Llama3:8b}
\newcommand{\smcalflow}{SMCalFlow}
\newcommand{\geoquery}{GeoQuery}
\newcommand{\mtop}{MTOP}
\newcommand{\tfive}{T5}
\newcommand{\tten}{T10}
\newcommand{\geoA}{D3}
\newcommand{\geoB}{D4}

\def\ztitle{Diverse  In-Context Example Selection After Decomposing Programs and Aligned Utterances Improves Semantic Parsing}

\title{\ztitle}

\author{%
  Mayank Kothyari\textsuperscript{1}\thanks{Corresponding authors: maykat2017@gmail.com, sunita@iitb.ac.in}, 
  Sunita Sarawagi\textsuperscript{1}\footnotemark[1], 
  Soumen Chakrabarti\textsuperscript{1} \\
  \textbf{Gaurav Arora\textsuperscript{2}, Srujana Merugu\textsuperscript{2}} \\
  \textsuperscript{1}Indian Institute of Technology Bombay \quad
  \textsuperscript{2}Amazon
}

\begin{document}
\maketitle


\begin{abstract}
LLMs are increasingly used as seq2seq translators from natural language utterances to structured programs, a process called semantic interpretation.
Unlike atomic labels or token sequences, programs are naturally represented as abstract syntax trees (ASTs).  Such structured representation raises novel issues related to the design and selection of in-context examples (ICEs) presented to the LLM.  We focus on \emph{decomposing} the pool of available ICE trees into fragments
 to minimize interference from irrelevant content and improve generalization on test instances.
Next, we propose how to use (additional invocations of) an LLM with prompted syntax constraints to automatically map the fragments to corresponding utterances.
Finally, we adapt and extend a recent method for diverse ICE selection to work with whole and fragmented ICE instances.  We evaluate our system, \sysname\footnote{The code and dataset for the paper are available at \url{https://github.com/iMayK/SCUD4ICL}}, on popular diverse semantic parsing benchmarks, showing visible accuracy gains from our proposed decomposed diverse demonstration method.  Benefits are particularly notable for smaller LLMs, ICE pools having larger labeled trees, and programs in lower resource languages.
\end{abstract}

\section{Introduction}
\label{sec:Intro}

Large language models (LLMs), being proficient program generators \citep{Yan2023CodeScopeAE}, are well-suited to solving semantic interpretation tasks: translating natural language \emph{utterances} $\vx_0$ (which could be questions or instructions) into executable \emph{code} $\vy_0$ (interchangeably called program, or query) in a structured language, such as Python, SQL or SPARQL, possibly even invoking libraries \citep{Hsieh2023LlmTool} such as PyTorch or Pandas~\citep{Ye2024DataFrameQA}. \todo{@SS check if code mention looks too dangerous} Here we characterize code using their abstract syntax trees (ASTs).

\begin{figure*}[t]
\centering
\adjustbox{max width=.8\hsize}{%
\includegraphics{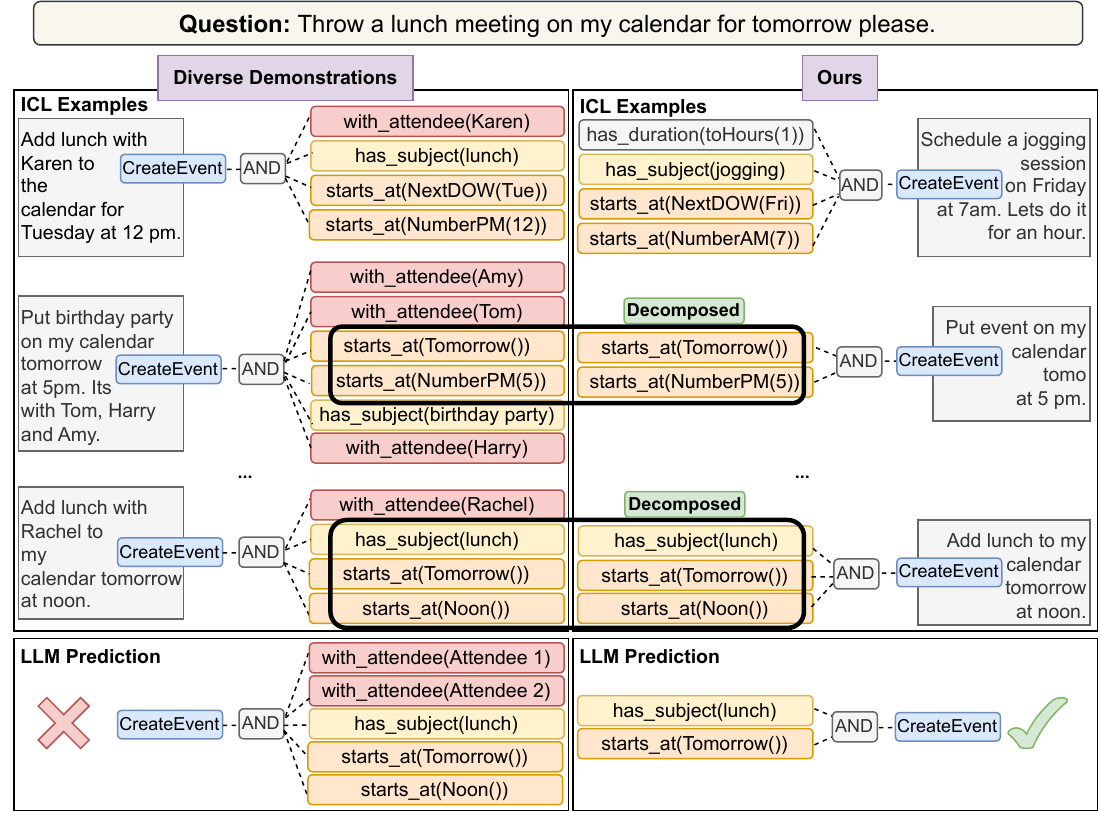}}
\caption{An example of how decomposed queries help avoid interference. On the left are three whole ICEs selected by an existing method.  On the right are \sysname's ICEs. Note that two of these are decompositions of training examples, after removing irrelevant clauses. Removing the irrelevant clauses reduces interference during ICL leading to a correct prediction from the LLM.}
\label{fig:intro_example}
\end{figure*}

Pretraining corpora of LLMs include diverse public schema, structured data and utterances, e.g., in the field of text2sql or knowledge graph question answering (KGQA).  When applied to less popular domains like calendar management \citep[SMCalFlow]{smcalflow} or inspecting geographical databases \citep[GeoQuery]{geoquery}, pretrained LLMs perform less impressively. This can be a serious impediment to exploiting LLMs for structured interpretation in settings where the schema and data are private, e.g., in case of enterprise data that were not part of the pretraining corpus.  Given the enormous size of the best LLMs, continued training or fine-tuning may be impractical for most users.

LLMs are also known to be effective in-context learners (ICL)~\citep{Brown2020LmFewShot, Lu2023EmergentInContext}.
This capability may be particularly beneficial \citep{LevyBB23} when in-context examples (ICEs) involve the same (possibly private) schema and task.
Formally, given test utterance $\vx_0$,  the input to the LLM includes $M$ in-context examples represented in the form $\bm{p}; \vx_1,\vy_1;\ldots;\vx_M,\vy_M;\vx_0$ with suitable delimiters. Here, $\bm{p}$ is an optional instruction prefix, and the LLM has to decode $\hat{\vy}_0$, the translation of~$\vx_0$.  
A common challenge for ICL is to select $S=\{(\vx_m,\vy_m):m\in[M]\}$\footnote{$ [M]$  refers to the set $\{1, 2, …M \}$}
from a larger corpus of instances to maximally assist the LLM, the guiding principles being:
\begin{enumerate*}[(1)]
\item $M$ should be small to reduce forward inference cost.
\item $\vx_m$ should be strongly related to $\vx_0$.
\item $S$ should be suitably diverse, in an attempt to provide adequate coverage of possible target ASTs.
\end{enumerate*}

Our point of departure is to bring a new desirable criterion into the above picture of ICL.  ASTs $\{\vy_m: m\in[M]\}$, as well as the target $\vy_0$ are complex structured outputs.  The ideal $\vy_0$ may have only \emph{partial overlap} with ASTs in the ICEs $\{\vy_m: m\in[M]\}$, but the non-overlapping parts may distract in-context learning. 
This motivates our first contribution:
select ICEs from not only `whole' ASTs, but also \emph{decompose} them into meaningful fragments (typically, subtrees) and make them available for the ICE subset selector.  This is in sharp contrast to all existing methods that choose whole examples for ICL demonstration.

The enhancement proposed above raises the issue that even if $(\vx_m,\vy_m)$ pairs are provided for complete utterances $\vx_m$ and corresponding complete ASTs $\vy_m$, and if the decomposition of $\vy_m$ into useful fragments $\vy_{m,k}$ were possible to automate, these AST fragments do not come with corresponding sub-utterances~$\vx_{m,k}$.  Our second contribution is to employ an LLM for this translation task, but with a twist that turns out to be critical: we instruct the LLM to regard the utterances $\vx_{m,k}$ to be generated as \emph{sub-utterances} of~$\vx_m$.

Our third contribution is, given $\vx_0$, to select, from the available pool of whole- and fragmented ICEs, a suitable subset to include in the LLM input. This is also a delicate step, because inclusion of ICEs irrelevant to $\vx_0$ (including cases with complete ASTs) have the potential \citep{liu2023lostmiddlelanguagemodels, Chen2023interference} to interfere with the generation of the correct parse~$\vy_0$.  We show examples of such in-context interference for a semantic parsing task in Figure~\ref{fig:intro_example}.  We extend a recent diverse demonstration method \citep{LevyBB23} to handle complete and decomposed ICEs seamlessly.

Our system, \sysname{} (\longname), incorporates all the three enhancements above, and will be released publicly on acceptance.

We present empirical evaluations with three popular diverse semantic parsing benchmarks:  SMCalFlow, GeoQuery, and MTOP. 
We explore the effects of diverse training sizes, train-test discrepancy, models, decomposition depths, and code languages.
Apart from consistent wins at semantic interpretation, we establish that fragmented ICE availability, coupled with our selection criterion, reduces harmful interference.
Benefits are particularly notable for smaller \todo{smaller?} LLMs, larger labeled trees, and lower resource languages.


\begin{figure*}[t]
\centering\adjustbox{max width=\hsize}{%
\includegraphics{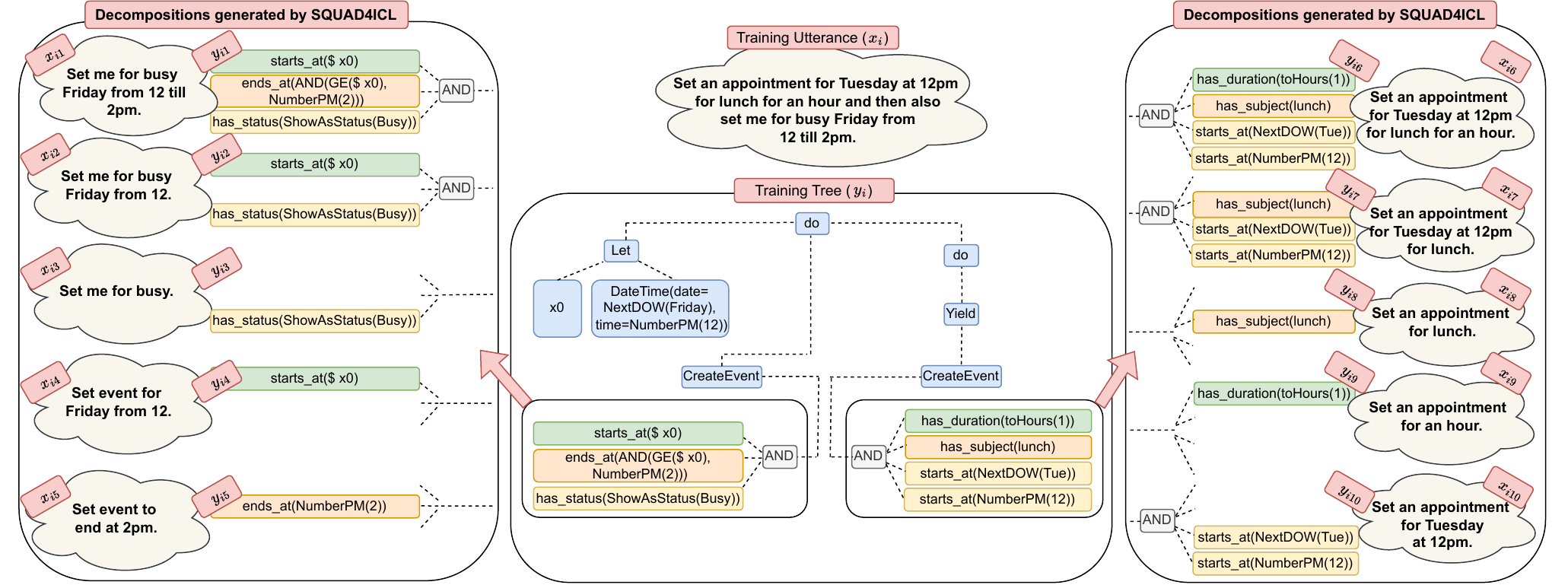}}
\caption{An example showing decomposition of a training instance by \sysname. A complex training utterance-tree pair ($\vx_i$, $\vy_i$) comprising of more than ten clauses is decomposed into ten subtrees of varying complexity.  The sub-utterances $\vx_{i,j}$ attached to each sub-tree $\vy_{i,j}$ are subsumed by $\vx_i$ while being fluent and relevant to the respective $\vy_{i,j}$. The \texttt{"Let"} clause, which defines $\vx_0$, is repeated in subqueries wherever needed, but we omit repetition in the figure to reduce clutter.}
\label{fig:example_decompositions}
\end{figure*}

\section{Related Work}
\label{sec:Rel}

Our current work on semantic parsing using LLMs with ICL leverages prior work on selection of ICEs and query decomposition techniques. 


\todo{SM: wanted to mention the "anonymization" of named entity tokens while matching for better generalization - wasn't sure of the papers to reference and left it out for now @mayank- is this important? did we do it?}

\paragraph{In-context example selection}
\todo{ SM:Shortening ideas: First few lines might be unnecessay, can remove hongjin reference or combine it with Gupta et al as an unsupervised strategy, TISE reference also is not really that important-just builds on Ye-CEIL, can remove Bogin et al and allude to the local structures as part of Levy-DD reference.}
It is crucial to select ICEs that are highly informative with respect to the test utterance. An intuitive approach is to select the top-$K$ utterances most similar to the test utterance~\citep{Liu2022, rubin2021learning}. 
However, this method often results in redundancy and limited  coverage. \citet{gupta-etal-2023-coverage} introduce an unsupervised set-selection approach to mitigate this problem. Similarly, \citet{hongjin2022selective} propose an unsupervised, graph-based strategy that combines similarity and diversity. However, their work focuses on reducing annotation of ICL examples, rather than query-specific selection.

\begin{figure*}[t]
\centering
\adjustbox{max width=\hsize}{%
\includegraphics{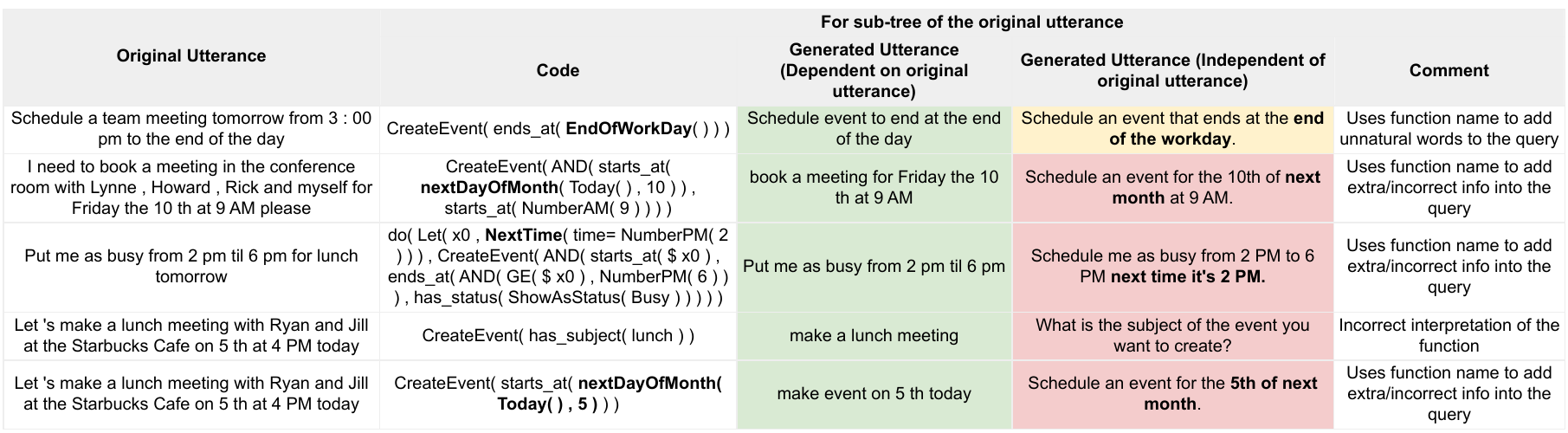}}
\caption{Examples showing how utterances generated by \sysname{} conditional on original training utterances are more fluent and natural than utterances generated when the LLM is not prompted to encourage subsumption.}
\label{fig:lang_pres_ablation}
\end{figure*}

More recently, \citet{Ye2023CEIL} have formulated ICL example selection as subset selection using Determinantal Point Processes (DPPs) \citep{kulesza2011learning}. This approach optimizes example selection through a contrastive learning objective that balances relevance and diversity, using a specially designed kernel. Building on this, \citet{fu-etal-2024-tise} propose TISE, a tripartite selection method that incorporates contextual relevance, event correlation, and example diversity.  For semantic parsing applications, where structure encodes vital information, \citet{bogin-etal-2024-leveraging,LevyBB23} propose generating  diverse demonstrations by
collectively covering the maximal number of local structures in the test utterance. 

All the above methods, however, treat both $\vx_m$ and $\vy_m$ as monolithic. When the selected ICEs contain irrelevant sub-parts, there is a high risk of the LLM being misled (as we shall establish). While our work builds on the diverse selection method of \citet{LevyBB23}, we first generate an enhanced pool of ICEs obtained by decomposition.  This enables the selection step to protect the LLM from interference from irrelevant fragments.



\paragraph{Example Decomposition} 
\todo{SM: wasn't sure if Eyal et al was required. Last para on the contrast also needs a review for correctness}
In complex question answering (QA) and semantic parsing,
the utterance is often decomposed into simpler sub-utterances that are mapped to sub-queries, whose responses are later assembled into the final answer.
\citet{liu2023hpe} propose a hybrid complex QA system performs top-down parsing of questions into tree-structured representations (referred to as H-expressions).
\citet{huang2023question} present a neural model that hierarchically decomposes complex questions into trees.  
\citet{shi2023exedec} propose a novel execution decomposition (ExeDec) strategy for tackling the broader problem of program synthesis by addressing key forms of compositional generalization. Their approach predicts execution subgoals to solve problems incrementally, guided by program execution at each step as well as carefully curated few shot LLM prompts.
For text2sql, \citet{eyal2023semantic} propose an intermediate Query Plan Language (QPL) more amenable to decomposition that SQL.
Lately, LLMs \citep{pourreza2024din, liu2023divide, wang2023mac} --- suitably prompted with ICEs and chain-of-thought (CoT) reasoning --- perform schema linking, query decomposition, SQL generation, and refinement for text2sql tasks.
%

These methods rely on \todo{@MK @SM static means what here?-SM:I meant the examples are not specific to query} \todo{MK: same set of examples for all test questions, i.e., not query specific} static few-shot examples and decomposition of the \emph{test} utterance, which can incur high latency and computational overhead. In contrast, our approach 
applies 
decomposition to the labeled \emph{training} instances, creating a larger ICE pool. We then rely on test-time selection to identify diverse and relevant ICEs (often choosing subtrees of the original ICEs). This approach avoids high latency and computational costs of test utterance decomposition during inference.

\paragraph{ICL issues affecting compositional generalization}
Given the broad applicability of ICL, numerous studies have explored how factors such as the model scale, token novelty, test query complexity, and demonstration placement influence overall performance.
\citet{qiu-etal-2022-evaluating} show that larger model sizes improve compositional generalization in semantic parsing, but heightens sensitivity to noise in ICEs, a trend also noted by \citet{shi2023largelanguagemodelseasily, wei2023largerlanguagemodelsincontext} and explained theoretically by \citet{shi2024largerlanguagemodelsincontext}. Building on this observation, \citet{wang-etal-2024-mdr} introduce a model-specific demonstration selection method tailored to LLM biases.
\citet{an-etal-2023-context} investigate the impact of similarity, diversity, and complexity in ICEs, concluding that optimal performance comes from examples that are structurally similar, diverse, and simple. They also emphasize the need for the ICEs to cover the linguistic structures of the test instance.
%
\todo{SM: Need to add a line to comment on  our findings on model size, complexity and rare words and how they add to or contrast with existing literature}

\section{Proposed method: \sysname}
\label{sec:Method}

We are given a training set $\train=\{(\vx_i,\vy_i):i=1,\ldots,N\}$, comprising  $N$ pairs of natural language utterances $\vx_i$ and their corresponding programs~$\vy_i$.  Our goal is to use $\train$ to provide an LLM acting as a semantic interpreter) a set $S$ of $M$ in-context examples for any test question $\vx_0$.  Unlike previous methods, where $S$ contained a subset of only the original examples in $\train$, our method \sysname\ seeks more focused sub-programs to reduce interference during in-context learning.  We achieve this in two steps: 
\begin{enumerate*}[(1)]
\item We perform a one-time decomposition of each training instance to create an augmented training pool $\AugD$, and 
\item Next, for each test utterance $\vx_0$, we select a focused, diverse set of examples from $\AugD$.
\end{enumerate*}
We describe these steps next.

\subsection{Instance decomposition}
\label{sec:Decomp}

Given an utterance-program pair ($\vx_i,\vy_i)$, our goal is to decompose $\vy_i$ into meaningful sub-programs $\vy_{i,1},\ldots,\vy_{i,K_i}$ and associate each sub-program $\vy_{i,k}$ with a corresponding utterance $\vx_{i,k}$ (which may have to be generated artificially).
We assume that $\vy_i$ is a semantic parse tree, and the tree structure naturally defines sub-programs corresponding to its subtrees.  
We show an example of a decomposition in Figure~\ref{fig:example_decompositions}.

\paragraph{Sub-utterance generation}
A baseline method to generate decompositions is to first extract sub-trees rooted at each internal node of the original tree $\vy_i$ following the grammar of the program.  Then for each subtree $\vy_{i,j}$, invoke an LLM to generate utterances for the subtrees, possibly using ICL for that intermediate task.
 We call this the \emph{``independent utterance decomposition''} method.

\begin{figure}
\begin{tcolorbox}[width=\hsize, colframe=black, colback=gray!5, boxrule=.1pt, left=0mm, right=0mm, top=0mm, bottom=0mm, label=qd_prompt]
\parskip 1.5ex \ttfamily \small
In this task, your goal is to decompose complex event scheduling queries into simpler, self-contained sub-queries.

Each sub-query should be represented as a key-value pair within a JSON object. 
The key is the sub-query, and the value is an object containing the corresponding Domain-Specific 
Language (DSL) code and any further decompositions. 

The primary objective is to ensure that the decompositions closely mirror the language used in the 
original query. This is to retain the natural language and idiosyncrasies of the user's input, 
which are crucial for understanding the context and intent of the query.

Each sub-query should be able to stand on its own, without relying on the context of the original 
query. This means that the sub-queries should be clear and unambiguous, even when viewed
independently of the original query.  
\end{tcolorbox}  
\caption{Instruction to LLM for subsumed utterance decomposition in SMCalFlow. These are followed by a few decomposition ICEs. Figure~\ref{fig:qd_ICL_1} shows a sample.}
\label{fig:prompt} 
\end{figure}

A limitation of the above method is that the synthetically generated utterance may not align with the style and language of human-generated utterance, and thus may not be useful demonstration for converting actual test utterances into programs.  Our key idea is to view $\vx_{i,j}$ as a \emph{sub-utterance} of $\vx_i$.  We harness LLMs for  generating (sub-utterance, sub-tree) pairs conditional on an $\vx_i,\vy_i$.
 We call this the \emph{``subsumed utterance decomposition''} method.
 The LLM is instructed to preserve as much of the original utterance in generating the sub-utterance, and also shown a few (manually created) examples of such decompositions.  In  Figure~\ref{fig:example_decompositions}, notice how each of generated sub-utterances are almost subsumed by the original utterance while being faithful to the given sub-tree.  In Figure~\ref{fig:lang_pres_ablation}, we contrast the utterance generated by our conditional prompting against independent generations. Note how the utterances generated independent for each sub-tree appear too verbatim and unnatural compared to our subsumed generation.
Figure~\ref{fig:prompt} shows the prompt used for such generation.

\begin{algorithm}[t]
\caption{SCUD4ICL Example Selection}
\label{alg:select}
\small
\begin{algorithmic}[1]
\Require
    \State $\AugD$: decomposed ICE pool $\{(\vx_{n,k}, \vy_{n,k}) : n \in [N], k \in [K_n]\}$
    \State $M$: budget for selected examples
    \State Test query: $\vx_0$
\Ensure Diverse and relevant selected examples $S$ 
    \State $S \gets \emptyset$                  \Comment{Selected examples}
    \State $T_{\text{covered}} \gets \emptyset$ \Comment{Covered test tokens}
    \State $T_{\text{seen}} \gets \emptyset$    \Comment{Seen anonymized templates}
    \While{$|S| < M$}
        \State $T_{\text{promising}} \gets \emptyset$
        \For{each token $x \in \vx_0$ such that $x \notin T_{\text{covered}}$}
            \State Identify candidates $(n,k)$ in $\AugD$ satisfying:
            \begin{enumerate}
                \item $(n,k) \notin S$
                \item No ancestor/descendant of $\vy_{n,k}$ is in $S$
                \item $\text{anonymized}(\vy_{n,k}) \notin T_{\text{seen}}$
                \item $x \in \vx_{n,k}$
            \end{enumerate}
            \If{valid candidates exist}
                \State $T_{\text{promising}} \gets T_{\text{promising}} \cup \{x\}$
            \EndIf
        \EndFor
        \If{$T_{\text{promising}} \subseteq T_{\text{covered}}$}
            \State $T_{\text{covered}} \gets \emptyset$ \Comment{Reset coverage}
            \State \textbf{continue}
        \EndIf
        \If{$T_{\text{promising}} = \emptyset$}
            \State \textbf{break}
        \EndIf
        \State $x \gets$ random token from $T_{\text{promising}}$
        \State $C \gets$ set of candidates in $\AugD$ for token $x$
        \State $(n^*, k^*) \gets \argmax_{(n,k) \in C} \operatorname{\href{https://en.wikipedia.org/wiki/Okapi_BM25}{BM25}}(\vx_0, \vx_{n,k})$
        \State $S \gets S \cup \{(n^*, k^*)\}$
        \State $T_{\text{seen}} \gets T_{\text{seen}} \cup \{\text{anonymized}(\vy_{n^*,k^*})\}$
        \State $T_{\text{covered}} \gets T_{\text{covered}} \cup \{x\}$
    \EndWhile
    \State \Return $S$
\end{algorithmic}
\end{algorithm}

\subsection{ICE selection from $\AugD$}

Let $\AugD$ denote the original dataset $\train$ augmented with generated decomposed pairs $\{(\vx_{n,k}, \vy_{n,k}): n=1,\ldots,N; k=1\ldots K_i\}$ pairs.  Given a test question $\vx$, we adapt the state of the art diverse decomposition algorithm of \citet{LevyBB23} to select examples from $\AugD$.  Let $M$ be the budgeted number of ICL examples to select.
We collect ICEs one by one.
At each step, a new candidate $(\vx_{n,k}, \vy_{n,k})$ is selected if it satisfies these criteria:
\begin{enumerate}
\item utterance or sub-utterance $\vx_{n,k}$ covers one or more hitherto uncovered token(s) in the test utterance $\vx_0$,
\item a descendant or ancestor node of the root of $\vy_{n,k}$ has not already been selected,
\item when the anonymized version of $\vy_{n,k}$ (i.e.,  with all entity names/values  replaced by a common token \texttt{ANON}) is not among the ones already seen, and 
\item $\vx_{n,k}$ is maximally similar to $\vx_0$ among candidates that satisfy the above criteria (similarity is measured using \href{https://en.wikipedia.org/wiki/Okapi_BM25}{BM25}).
\end{enumerate}
Algorithm~\ref{alg:select} presents our ICE selection strategy.
As can be seen, various tweaks to diverse decomposition \citep{LevyBB23} are needed to accommodate candidates corresponding to program fragments that may have structured relations between them, including matching against canonical anonymized structural sketches of ICE sub-programs already accepted into~$S$. Algorithm~\ref{alg:select} can be viewed as solving an optimization problem  balancing relevance to the test query $\vx_0$  and diversity within the selected set $S$ constrained by a budget $M$ using an efficient greedy algorithm. Refer to Appendix  \ref{appendix:optimization} for a more detailed discussion.

\begin{table*}[ht]
\centering\adjustbox{max width=.8\hsize}{
\begin{tabular}{|l|l|c|c|c|c||c|c|c|c|}
\hline
\multirow{2}{*}{\textbf{LLM}} & \multirow{2}{*}{\textbf{ICL type}} & \multicolumn{4}{c|}{\textbf{SMCalFlow}} & \multicolumn{4}{c|}{\textbf{GeoQuery}} \\ \cline{3-10}
                                &                                  & EN & HI & FR & RU & length & i.i.d & template & tmcd  \\ \hline
\multirow{2}{*}{\chatgpt
} & \baseline      & 42.4    & 43.4    & 40.1    & 48.7    & 50.9 & 77.2 & 86.5 & 70.9  \\ 
            
                    & \sysname & \textbf{45.8 }   & \textbf{50.2 }   & \textbf{46.0 }   & \textbf{49.5}   & \textbf{59.2} & \textbf{81.9} & \textbf{87.7} & \textbf{82.0}  \\ \hline
\multirow{2}{*}{\gpt}        & \baseline      & 54.1    & 49.6    & 53.5    & 54.5  & 73.6 & 86.3 & 91.6 & 80.7  \\ 
                                & \sysname & \textbf{54.2}    & \textbf{51.8 }   & \textbf{54.3}    & \textbf{55.8} & \textbf{81.0} & \textbf{86.9} & 89.8 & \textbf{85.4}   \\ \hline
\multirow{2}{*}{\mistral}     & \baseline      & 45.8    & 32.7    & 43.5    & 36.5  & 32.3 & 58.9 & 62.3 & 49.4   \\ 
                                & \sysname & \textbf{46.6 }   & \textbf{37.2}    & \textbf{46.0 }   &\textbf{ 39.8 }  & \textbf{50.6 }& \textbf{67.7 }& \textbf{74.3} & \textbf{73.9}  \\ \hline
\multirow{2}{*}{\llama}      & \baseline      & 48.2    & 28.5    & 43.5    & 39.3  & 35.6 & 55.8 & 63.4 & 50.9  \\ 
                                & \sysname & 46.4    & \textbf{34.2}    & \textbf{43.7 }   & \textbf{40.2  }  & \textbf{44.4} & 52.9 & 58.9 & 48.4 \\ \hline

\end{tabular}}
\caption{Execution accuracy on two datasets: (1) SMCalFlow with training split $\train=$ \tfive\ with programs in four different languages En,Hi,Fr,Ru. and (2) GeoQuery with training split $\train=$ \geoA\ and various test splits.  All results are  average of three runs and with $M=5$. Observe that \sysname\ provides much higher accuracy than \baseline\ in most cases, and the gains are higher for smaller LMs.}
\label{tab:overall}
\end{table*}

\begin{table}[th]
\centering\adjustbox{max width=.9\hsize}{
\begin{tabular}{|l|l|c|c|c|c|}
\hline
\multirow{2}{*}{\textbf{LLM}} & \multirow{2}{*}{\textbf{ICL type}} & \multicolumn{2}{c|}{EN} & \multicolumn{2}{c|}{FR} \\ \cline{3-6} 
                              &                                & random  & full & random  & full \\ \hline
\multicolumn{6}{|c|}{\textbf{M=5}} \\ \hline
\multirow{2}{*}{\mistral}      & CoverLS                       & 25.0           & 22.5        & 17.3           & 18.2 \\
                              & SCUD4ICL                      & \textbf{27.3}           & \textbf{22.8}        & \textbf{19.7}           & \textbf{19.6} \\ \hline
\multirow{2}{*}{\llama}      & CoverLS                       & 17.3           & 14.9        & 11.0           & 11.3 \\
                              & SCUD4ICL                      & \textbf{18.7}           & \textbf{16.5}        & \textbf{11.3}           & \textbf{12.1} \\ \hline
\multicolumn{6}{|c|}{\textbf{M=10}} \\ \hline
\multirow{2}{*}{\mistral}      & CoverLS                       & 29.7           & 24.0        & 23.0           & 21.1 \\
                              & SCUD4ICL                      & \textbf{32.0}           & \textbf{26.4}        & \textbf{25.3}           & \textbf{22.9} \\ \hline
\multirow{2}{*}{\llama}      & CoverLS                       & 17.3           & 16.1        & 13.7           & 13.0 \\
                              & SCUD4ICL                      & \textbf{23.7}           & \textbf{20.8}        & \textbf{15.3}          & \textbf{14.6} \\ \hline
\end{tabular}}
\caption{Exact match accuracy on MTOP for EN and FR for $M=5$ and $M=10$. Results on random set are averaged over three runs.}
\label{tab:mtop}
\end{table}

\begin{table}[ht]
    \centering\adjustbox{max width=.8\hsize}{
    \begin{tabular}{|l|l|c|c|}
        \hline
        \textbf{LLM}          & \textbf{ICL type} & EN   & HI   \\ \hline
        \chatgpt              & DPP              & 19.3          & 24.6          \\ 
                              & TOPK             & 24.9          & 25.5          \\ 
                              & \baseline        & 42.4          & 43.2          \\ 
                              & \sysname         & \textbf{45.7}          & \textbf{50.5}          \\ \hline
        \mistral              & DPP              & 22.9          & 23.4          \\ 
                              & TOPK             & 25.3          & 24.0          \\ 
                              & \baseline        & 45.4          & 32.1          \\ 
                              & \sysname         & \textbf{46.0}          & \textbf{36.7}          \\ \hline
        \llama                & DPP              & 27.0          & 22.5          \\ 
                              & TOPK             & 27.9          & 22.0          \\ 
                              & \baseline        & \textbf{48.0}          & 27.6          \\ 
                              & \sysname         & 45.3          & \textbf{34.1}          \\ \hline
    \end{tabular}}
    \caption{\sysname\ vs. other baselines on SMCalFlow (training split: $\train=$ \tfive, $M=5$).}
    \label{tab:sm_otherbaselines}
\end{table}

\section{Experimental Setup}
\label{sec:data}

We experiment with two semantic parsing datasets.

\paragraph{\smcalflow~\cite{smcalflow}}
SMCalFlow-CS is a  dataset of approximately 25 thousand human-generated utterances about calendar management.   To mimic realistic settings of limited labeled data, we created two different training sets, $\tfive$ and $\tten$, as follows.  For each test sample, we selected 5 (respectively, 10) closest samples  on utterance-level cosine similarity.  We retain those of depth greater than three. This yielded 71 unique instances in $\tfive$, and 116 in $\tten$.  After decomposition with our method we obtained 296 and 473 non-leaf decompositions which we treat as the augmented training set~$\AugD$.  The default train pool is $\tfive$. In these experiments we were concerned if the recent LLMs were already trained or fine-tuned on these datasets (data contamination).  To partially mitigate this concern, we created three variants of the data, where English names in clauses are replaced by their counterpart in three other languages:
Hindi, French, and Russian. The utterance $\vx_i$ stays in English, only in $\vy_i$ the names of clauses are changed using a mapping as shown in Figure~\ref{schema:en}, ~\ref{schema:hi} and ~\ref{schema:fr}
in the Appendix. We will soon see that changing clause names impacts performance, particularly in recent smaller LLMs.


\paragraph{\geoquery~\cite{geoquery}} is a dataset of 880 user utterances seeking geographical information, such as locations of rivers and cities. 

\noindent\emph{\bfseries Test splits:}
For test data we use the \textit{iid} (standard) and compositional splits created by ~\citet{shaw-etal-2021-compositional} as follows:
\begin{enumerate*}[(1)]
\item \textit{template} split where programs output templates instead of grounded values for arguments 
\cite{finegan-dollak-etal-2018-improving};
\item \textit{TMCD} split, with divergent distributions of compounds in training and test sets~\cite{keysers2019measuring}; and
\item \textit{length} split, where test sequences are longer than training ones.     
\end{enumerate*}
As in prior work, we average results over three TMCD and template splits to reduce variance from small dataset size. 

\noindent\emph{\bfseries Train splits:}
Out of the available labeled dataset, we considered two subsets for defining the candidate labeled pool $\train$:
\begin{enumerate*}[(1)]
\item \geoA, comprising of trees of depth $\ge$ 3 (default), and
\item \geoB, comprising of trees of depth $\ge 4$.
\end{enumerate*}
For the \textit{length} split, since the test splits includes all the larger trees, the train split had only 8 trees left of depth~4.  So we do not consider this train-test split.  Across the other splits, the size of $\geoA$ is roughly four times the size of~$\geoB$.  Exact counts appear in Table~\ref{tab:smcal_geo_var_m_var_t}.

\paragraph{\mtop~\cite{li-etal-2021-mtop}} is a task-oriented dialogue dataset that maps user commands to complex, nested queries spanning 11 domains. 
We utilize the English and French subsets in our experiments. 
The original training sets include 15,667 English and 11,814 French utterances; 
we retain instances with query depth of at least 5, reducing them to 416 English and 211 French examples.
For controlled cross-lingual evaluation, we further filter the data to include only instances with corresponding queries in the other language, yielding a parallel training set of 108 examples. The test set comprises 1,713 parallel examples per language.

\paragraph{Models and Prompts}
We evaluate our method and baselines on the following LLMs: GPT\,3.5\footnote{\url{https://platform.openai.com/docs/models/gpt-3-5-turbo}}, GPT-4o\footnote{\url{https://openai.com/index/hello-gpt-4o/}}, Mistral\footnote{\url{https://mistral.ai/news/announcing-mistral-7b/}}-7b v0.3, and LLama3--8b~\cite{dubey2024llama3herdmodels}.
%
The prompt used for decomposition is shown in Figure~\ref{fig:prompt}. We use GPT4\_0125 to get the decompositions.  The prompt used for generating semantic parses is shown in Figure~\ref{prompt:semparse:en}.

\begin{table*}[t]
\adjustbox{max width=\hsize}{%
\tabcolsep 2pt
\begin{tabular}{|ll|cccccccc|cccccccc|}
\hline
\rowcolor[HTML]{C0C0C0} 
\multicolumn{1}{|l|}{\cellcolor[HTML]{EFEFEF}}                                      & \cellcolor[HTML]{EFEFEF}                                    & \multicolumn{8}{c|}{\cellcolor[HTML]{C0C0C0}\textbf{SMCalflow}}                                                                                                                                                                                                                                                                                                                                                                            & \multicolumn{8}{c|}{\cellcolor[HTML]{C0C0C0}\textbf{Geoquery}}                                                                                                                                                                                                                                                                                                                                                                                                                           \\ \cline{3-18} 
\rowcolor[HTML]{EFEFEF} 
\multicolumn{1}{|l|}{\cellcolor[HTML]{EFEFEF}}                                      & \cellcolor[HTML]{EFEFEF}                                    & \multicolumn{4}{c|}{\cellcolor[HTML]{EFEFEF}\textbf{M=5}}                                                                                                                                                                                 & \multicolumn{4}{c|}{\cellcolor[HTML]{EFEFEF}\textbf{M=10}}                                                                                                                                     & \multicolumn{4}{c|}{\cellcolor[HTML]{EFEFEF}\textbf{M=5}}                                                                                                                                                                                                         & \multicolumn{4}{c|}{\cellcolor[HTML]{EFEFEF}\textbf{M=10}}                                                                                                                                                           \\ \cline{3-18} 
\rowcolor[HTML]{EFEFEF} 
\multicolumn{1}{|l|}{\cellcolor[HTML]{EFEFEF}}                                      & \cellcolor[HTML]{EFEFEF}                                    & \multicolumn{4}{c|}{\cellcolor[HTML]{EFEFEF}\textbf{Languages}}                                                                                                                                                                           & \multicolumn{4}{c|}{\cellcolor[HTML]{EFEFEF}\textbf{Languages}}                                                                                                                                & \multicolumn{4}{c|}{\cellcolor[HTML]{EFEFEF}\textbf{Test Splits}}                                                                                                                                                                                                 & \multicolumn{4}{c|}{\cellcolor[HTML]{EFEFEF}\textbf{Test Splits}}                                                                                                                                                    \\ \cline{3-18} 
\rowcolor[HTML]{EFEFEF} 
\multicolumn{1}{|l|}{\multirow{-4}{*}{\cellcolor[HTML]{EFEFEF}\textbf{LLM}}}        & \multirow{-4}{*}{\cellcolor[HTML]{EFEFEF}\textbf{ICL type}} & \multicolumn{1}{c|}{\cellcolor[HTML]{EFEFEF}\textbf{EN}} & \multicolumn{1}{c|}{\cellcolor[HTML]{EFEFEF}\textbf{HI}} & \multicolumn{1}{c|}{\cellcolor[HTML]{EFEFEF}\textbf{FR}} & \multicolumn{1}{c|}{\cellcolor[HTML]{EFEFEF}\textbf{RU}} & \multicolumn{1}{c|}{\cellcolor[HTML]{EFEFEF}\textbf{EN}} & \multicolumn{1}{c|}{\cellcolor[HTML]{EFEFEF}\textbf{HI}} & \multicolumn{1}{c|}{\cellcolor[HTML]{EFEFEF}\textbf{FR}} & \textbf{RU}   & \multicolumn{1}{c|}{\cellcolor[HTML]{EFEFEF}\textbf{length}} & \multicolumn{1}{c|}{\cellcolor[HTML]{EFEFEF}\textbf{iid}} & \multicolumn{1}{c|}{\cellcolor[HTML]{EFEFEF}\textbf{template}} & \multicolumn{1}{c|}{\cellcolor[HTML]{EFEFEF}\textbf{tmcd}} & \multicolumn{1}{c|}{\cellcolor[HTML]{EFEFEF}\textbf{length}} & \multicolumn{1}{c|}{\cellcolor[HTML]{EFEFEF}\textbf{iid}} & \multicolumn{1}{c|}{\cellcolor[HTML]{EFEFEF}\textbf{template}} & \textbf{tmcd} \\ \hline
\rowcolor[HTML]{EFEFEF} 
\multicolumn{2}{|l|}{\cellcolor[HTML]{EFEFEF}\textbf{Train Split}}                                                                                & \multicolumn{8}{c|}{\cellcolor[HTML]{EFEFEF}$\train=$ \tfive}                                                                                                                                                                                                                                                                                                                                                                               & \multicolumn{8}{c|}{\cellcolor[HTML]{EFEFEF}$\train=$ D4}                                                                                                                                                                                                                                                                                                                                                                                                                             \\ \hline
\rowcolor[HTML]{EFEFEF} 
\multicolumn{2}{|l|}{\cellcolor[HTML]{EFEFEF}\textbf{No. of instances}}                                                                           & \multicolumn{8}{c|}{\cellcolor[HTML]{EFEFEF}71/296}                                                                                                                                                                                                                                                                                                                                                                               & \multicolumn{1}{c|}{\cellcolor[HTML]{EFEFEF}2/0}    & \multicolumn{1}{c|}{\cellcolor[HTML]{EFEFEF}54/176}   & \multicolumn{1}{c|}{\cellcolor[HTML]{EFEFEF}35/120}      & \multicolumn{1}{c|}{\cellcolor[HTML]{EFEFEF}34/107}  & \multicolumn{1}{c|}{\cellcolor[HTML]{EFEFEF}2/0}    & \multicolumn{1}{c|}{\cellcolor[HTML]{EFEFEF}54/176}   & \multicolumn{1}{c|}{\cellcolor[HTML]{EFEFEF}35/120}      & 34/107  \\ \hline
\multicolumn{1}{|l|}{\cellcolor[HTML]{EFEFEF}\textbf{\chatgpt}}                       & \cellcolor[HTML]{EFEFEF}\textbf{\baseline}                    & \multicolumn{1}{c|}{42.4}                                & \multicolumn{1}{c|}{43.4}                                & \multicolumn{1}{c|}{40.1}                                & \multicolumn{1}{c|}{48.7}                                & \multicolumn{1}{c|}{42.6}                                & \multicolumn{1}{c|}{50.1}                                & \multicolumn{1}{c|}{41.4}                                & 47.1          & \multicolumn{1}{c|}{-}                                       & \multicolumn{1}{c|}{64.4}                                      & \multicolumn{1}{c|}{69.9}                                         & \multicolumn{1}{c|}{63.4}                                     & \multicolumn{1}{c|}{-}                                       & \multicolumn{1}{c|}{71.5}                                      & \multicolumn{1}{c|}{73.7}                                         & 67.3             \\ \cline{1-1}
\multicolumn{1}{|l|}{\cellcolor[HTML]{EFEFEF}\textbf{}}                             & \cellcolor[HTML]{EFEFEF}\textbf{\sysname}                   & \multicolumn{1}{c|}{\textbf{45.8}}                       & \multicolumn{1}{c|}{\textbf{50.2}}                       & \multicolumn{1}{c|}{\textbf{46.0}}                       & \multicolumn{1}{c|}{\textbf{49.5}}                       & \multicolumn{1}{c|}{\textbf{50.8}}                       & \multicolumn{1}{c|}{\textbf{55.8}}                       & \multicolumn{1}{c|}{\textbf{50.9}}                       & \textbf{53.3} & \multicolumn{1}{c|}{-}                                       & \multicolumn{1}{c|}{\textbf{69.4}}                             & \multicolumn{1}{c|}{\textbf{71.3}}                                & \multicolumn{1}{c|}{\textbf{66.0}}                            & \multicolumn{1}{c|}{-}                                       & \multicolumn{1}{c|}{\textbf{78.6}}                             & \multicolumn{1}{c|}{\textbf{81.3}}                                & \textbf{68.5}    \\ \hline
\multicolumn{1}{|l|}{\cellcolor[HTML]{EFEFEF}}                                      & \cellcolor[HTML]{EFEFEF}\textbf{\baseline}                    & \multicolumn{1}{c|}{54.1}                                & \multicolumn{1}{c|}{49.6}                                & \multicolumn{1}{c|}{53.5}                                & \multicolumn{1}{c|}{54.5}                                & \multicolumn{1}{c|}{\textbf{67.6}}                       & \multicolumn{1}{c|}{\textbf{64.6}}                       & \multicolumn{1}{c|}{\textbf{67.2}}                       & \textbf{67.0} & \multicolumn{1}{c|}{-}                                       & \multicolumn{1}{c|}{74.3}                                      & \multicolumn{1}{c|}{\textbf{85.5}}                                & \multicolumn{1}{c|}{\textbf{72.5}}                            & \multicolumn{1}{c|}{-}                                       & \multicolumn{1}{c|}{81.0}                                      & \multicolumn{1}{c|}{\textbf{87.0}}                                & 75.9             \\
\multicolumn{1}{|l|}{\multirow{-2}{*}{\cellcolor[HTML]{EFEFEF}\textbf{\gpt}}}      & \cellcolor[HTML]{EFEFEF}\textbf{\sysname}                   & \multicolumn{1}{c|}{\textbf{54.2}}                       & \multicolumn{1}{c|}{\textbf{51.8}}                       & \multicolumn{1}{c|}{\textbf{54.3}}                       & \multicolumn{1}{c|}{\textbf{55.8}}                       & \multicolumn{1}{c|}{65.4}                                & \multicolumn{1}{c|}{64.0}                                & \multicolumn{1}{c|}{65.9}                                & 66.4          & \multicolumn{1}{c|}{-}                                       & \multicolumn{1}{c|}{\textbf{76.8}}                             & \multicolumn{1}{c|}{81.4}                                         & \multicolumn{1}{c|}{71.9}                                     & \multicolumn{1}{c|}{-}                                       & \multicolumn{1}{c|}{\textbf{84.0}}                             & \multicolumn{1}{c|}{85.9}                                         & \textbf{77.1}    \\ \hline
\multicolumn{1}{|l|}{\cellcolor[HTML]{EFEFEF}}                                      & \cellcolor[HTML]{EFEFEF}\textbf{\baseline}                    & \multicolumn{1}{c|}{45.8}                                & \multicolumn{1}{c|}{32.7}                                & \multicolumn{1}{c|}{43.5}                                & \multicolumn{1}{c|}{36.5}                                & \multicolumn{1}{c|}{\textbf{53.7}}                       & \multicolumn{1}{c|}{35.3}                                & \multicolumn{1}{c|}{49.0}                                & 43.6          & \multicolumn{1}{c|}{-}                                       & \multicolumn{1}{c|}{45.7}                                      & \multicolumn{1}{c|}{\textbf{51.5}}                                & \multicolumn{1}{c|}{\textbf{45.9}}                            & \multicolumn{1}{c|}{-}                                       & \multicolumn{1}{c|}{54.2}                                      & \multicolumn{1}{c|}{54.8}                                         & 47.4             \\
\multicolumn{1}{|l|}{\multirow{-2}{*}{\cellcolor[HTML]{EFEFEF}\textbf{\mistral}}} & \cellcolor[HTML]{EFEFEF}\textbf{\sysname}                   & \multicolumn{1}{c|}{\textbf{46.6}}                       & \multicolumn{1}{c|}{\textbf{37.2}}                       & \multicolumn{1}{c|}{\textbf{46.0}}                       & \multicolumn{1}{c|}{\textbf{39.8}}                       & \multicolumn{1}{c|}{52.3}                                & \multicolumn{1}{c|}{\textbf{40.9}}                       & \multicolumn{1}{c|}{\textbf{52.8}}                       & \textbf{45.7} & \multicolumn{1}{c|}{-}                                       & \multicolumn{1}{c|}{\textbf{51.8}}                             & \multicolumn{1}{c|}{51.1}                                         & \multicolumn{1}{c|}{43.8}                                     & \multicolumn{1}{c|}{-}                                       & \multicolumn{1}{c|}{\textbf{58.2}}                             & \multicolumn{1}{c|}{\textbf{59.1}}                                & \textbf{47.8}    \\ \hline
\multicolumn{1}{|l|}{\cellcolor[HTML]{EFEFEF}}                                      & \cellcolor[HTML]{EFEFEF}\textbf{\baseline}                    & \multicolumn{1}{c|}{\textbf{48.2}}                       & \multicolumn{1}{c|}{28.5}                                & \multicolumn{1}{c|}{43.5}                                & \multicolumn{1}{c|}{39.3}                                & \multicolumn{1}{c|}{\textbf{57.9}}                       & \multicolumn{1}{c|}{33.8}                                & \multicolumn{1}{c|}{52.1}                                & 46.8          & \multicolumn{1}{c|}{-}                                       & \multicolumn{1}{c|}{36.7}                                      & \multicolumn{1}{c|}{41.9}                                         & \multicolumn{1}{c|}{39.4}                                     & \multicolumn{1}{c|}{-}                                       & \multicolumn{1}{c|}{51.2}                                      & \multicolumn{1}{c|}{48.2}                                         & 45.9             \\
\multicolumn{1}{|l|}{\multirow{-2}{*}{\cellcolor[HTML]{EFEFEF}\textbf{\llama}}}  & \cellcolor[HTML]{EFEFEF}\textbf{\sysname}                   & \multicolumn{1}{c|}{46.4}                                & \multicolumn{1}{c|}{\textbf{34.2}}                       & \multicolumn{1}{c|}{\textbf{43.7}}                       & \multicolumn{1}{c|}{\textbf{40.2}}                       & \multicolumn{1}{c|}{57.3}                                & \multicolumn{1}{c|}{\textbf{39.6}}                       & \multicolumn{1}{c|}{\textbf{52.7}}                       & \textbf{50.2} & \multicolumn{1}{c|}{-}                                       & \multicolumn{1}{c|}{\textbf{50.0}}                             & \multicolumn{1}{c|}{\textbf{47.6}}                                & \multicolumn{1}{c|}{\textbf{43.1}}                            & \multicolumn{1}{c|}{-}                                       & \multicolumn{1}{c|}{\textbf{61.9}}                             & \multicolumn{1}{c|}{\textbf{61.6}}                                & \textbf{51.8}    \\ \hline
\rowcolor[HTML]{EFEFEF} 
\multicolumn{2}{|l|}{\cellcolor[HTML]{EFEFEF}\textbf{Train Split}}                                                                                & \multicolumn{8}{c|}{\cellcolor[HTML]{EFEFEF}$\train=$ \tten}                                                                                                                                                                                                                                                                                                                                                                              & \multicolumn{8}{c|}{\cellcolor[HTML]{EFEFEF}$\train=$ D3}                                                                                                                                                                                                                                                                                                                                                                                                                             \\ \hline
\rowcolor[HTML]{EFEFEF} 
\multicolumn{2}{|l|}{\cellcolor[HTML]{EFEFEF}\textbf{No. of instances}}                                                                           & \multicolumn{8}{c|}{\cellcolor[HTML]{EFEFEF}116/473}                                                                                                                                                                                                                                                                                                                                                                             & \multicolumn{1}{c|}{\cellcolor[HTML]{EFEFEF}64/202} & \multicolumn{1}{c|}{\cellcolor[HTML]{EFEFEF}208/701}  & \multicolumn{1}{c|}{\cellcolor[HTML]{EFEFEF}147/576}     & \multicolumn{1}{c|}{\cellcolor[HTML]{EFEFEF}131/646} & \multicolumn{1}{c|}{\cellcolor[HTML]{EFEFEF}64/202} & \multicolumn{1}{c|}{\cellcolor[HTML]{EFEFEF}208/701}  & \multicolumn{1}{c|}{\cellcolor[HTML]{EFEFEF}147/576}     & 131/646 \\ \hline
\multicolumn{1}{|l|}{\cellcolor[HTML]{EFEFEF}\textbf{\chatgpt}}                       & \cellcolor[HTML]{EFEFEF}\textbf{\baseline}                    & \multicolumn{1}{c|}{44.9}                                & \multicolumn{1}{c|}{49.3}                                & \multicolumn{1}{c|}{41.3}                                & \multicolumn{1}{c|}{48.7}                                & \multicolumn{1}{c|}{44.1}                                & \multicolumn{1}{c|}{53.1}                                & \multicolumn{1}{c|}{44.5}                                & 51.5          & \multicolumn{1}{c|}{50.9}                                    & \multicolumn{1}{c|}{77.2}                                      & \multicolumn{1}{c|}{86.5}                                         & \multicolumn{1}{c|}{70.9}                                     & \multicolumn{1}{c|}{54.5}                                    & \multicolumn{1}{c|}{80.5}                                      & \multicolumn{1}{c|}{\textbf{88.3}}                                & 73.4             \\ \cline{1-1}
\multicolumn{1}{|l|}{\cellcolor[HTML]{EFEFEF}\textbf{}}                             & \cellcolor[HTML]{EFEFEF}\textbf{\sysname}                   & \multicolumn{1}{c|}{\textbf{50.7}}                       & \multicolumn{1}{c|}{\textbf{54.8}}                       & \multicolumn{1}{c|}{\textbf{52.5}}                       & \multicolumn{1}{c|}{\textbf{55.0}}                       & \multicolumn{1}{c|}{\textbf{54.3}}                       & \multicolumn{1}{c|}{\textbf{58.6}}                       & \multicolumn{1}{c|}{\textbf{53.8}}                       & \textbf{58.1} & \multicolumn{1}{c|}{\textbf{59.2}}                           & \multicolumn{1}{c|}{\textbf{81.9}}                             & \multicolumn{1}{c|}{\textbf{87.7}}                                & \multicolumn{1}{c|}{\textbf{82.0}}                            & \multicolumn{1}{c|}{\textbf{64.4}}                           & \multicolumn{1}{c|}{\textbf{84.6}}                             & \multicolumn{1}{c|}{88.1}                                         & \textbf{86.3}    \\ \hline
\multicolumn{1}{|l|}{\cellcolor[HTML]{EFEFEF}}                                      & \cellcolor[HTML]{EFEFEF}\textbf{\baseline}                    & \multicolumn{1}{c|}{56.9}                                & \multicolumn{1}{c|}{\textbf{56.6}}                       & \multicolumn{1}{c|}{\textbf{58.7}}                       & \multicolumn{1}{c|}{\textbf{62.0}}                       & \multicolumn{1}{c|}{\textbf{70.3}}                       & \multicolumn{1}{c|}{\textbf{69.3}}                       & \multicolumn{1}{c|}{67.6}                                & \textbf{70.9} & \multicolumn{1}{c|}{73.6}                                    & \multicolumn{1}{c|}{86.3}                                      & \multicolumn{1}{c|}{\textbf{91.6}}                                & \multicolumn{1}{c|}{80.7}                                     & \multicolumn{1}{c|}{81.2}                                    & \multicolumn{1}{c|}{88.4}                                      & \multicolumn{1}{c|}{\textbf{93.8}}                                & 84.1             \\
\multicolumn{1}{|l|}{\multirow{-2}{*}{\cellcolor[HTML]{EFEFEF}\textbf{\gpt}}}      & \cellcolor[HTML]{EFEFEF}\textbf{\sysname}                   & \multicolumn{1}{c|}{\textbf{58.1}}                       & \multicolumn{1}{c|}{56.4}                                & \multicolumn{1}{c|}{58.0}                                & \multicolumn{1}{c|}{61.1}                                & \multicolumn{1}{c|}{69.0}                                & \multicolumn{1}{c|}{67.0}                                & \multicolumn{1}{c|}{\textbf{67.7}}                       & 69.9          & \multicolumn{1}{c|}{\textbf{81.0}}                           & \multicolumn{1}{c|}{\textbf{86.9}}                             & \multicolumn{1}{c|}{89.8}                                         & \multicolumn{1}{c|}{\textbf{85.4}}                            & \multicolumn{1}{c|}{\textbf{82.4}}                           & \multicolumn{1}{c|}{\textbf{89.0}}                             & \multicolumn{1}{c|}{91.5}                                         & \textbf{90.8}    \\ \hline
\multicolumn{1}{|l|}{\cellcolor[HTML]{EFEFEF}}                                      & \cellcolor[HTML]{EFEFEF}\textbf{\baseline}                    & \multicolumn{1}{c|}{\textbf{50.2}}                       & \multicolumn{1}{c|}{38.4}                                & \multicolumn{1}{c|}{\textbf{48.5}}                       & \multicolumn{1}{c|}{\textbf{42.0}}                       & \multicolumn{1}{c|}{55.1}                                & \multicolumn{1}{c|}{41.7}                                & \multicolumn{1}{c|}{52.4}                                & 48.4          & \multicolumn{1}{c|}{32.3}                                    & \multicolumn{1}{c|}{58.9}                                      & \multicolumn{1}{c|}{62.3}                                         & \multicolumn{1}{c|}{49.4}                                     & \multicolumn{1}{c|}{37.2}                                    & \multicolumn{1}{c|}{62.5}                                      & \multicolumn{1}{c|}{69.5}                                         & 56.6             \\
\multicolumn{1}{|l|}{\multirow{-2}{*}{\cellcolor[HTML]{EFEFEF}\textbf{\mistral}}} & \cellcolor[HTML]{EFEFEF}\textbf{\sysname}                   & \multicolumn{1}{c|}{48.1}                                & \multicolumn{1}{c|}{\textbf{39.2}}                       & \multicolumn{1}{c|}{47.2}                                & \multicolumn{1}{c|}{41.2}                                & \multicolumn{1}{c|}{\textbf{55.9}}                       & \multicolumn{1}{c|}{\textbf{44.9}}                       & \multicolumn{1}{c|}{\textbf{54.3}}                       & \textbf{49.9} & \multicolumn{1}{c|}{\textbf{50.6}}                           & \multicolumn{1}{c|}{\textbf{67.7}}                             & \multicolumn{1}{c|}{\textbf{74.3}}                                & \multicolumn{1}{c|}{\textbf{73.9}}                            & \multicolumn{1}{c|}{\textbf{57.9}}                           & \multicolumn{1}{c|}{\textbf{74.5}}                             & \multicolumn{1}{c|}{\textbf{82.3}}                                & \textbf{77.3}    \\ \hline
\multicolumn{1}{|l|}{\cellcolor[HTML]{EFEFEF}}                                      & \cellcolor[HTML]{EFEFEF}\textbf{\baseline}                    & \multicolumn{1}{c|}{\textbf{51.2}}                       & \multicolumn{1}{c|}{35.3}                                & \multicolumn{1}{c|}{46.4}                                & \multicolumn{1}{c|}{\textbf{44.3}}                       & \multicolumn{1}{c|}{\textbf{59.1}}                       & \multicolumn{1}{c|}{37.1}                                & \multicolumn{1}{c|}{52.8}                                & 52.1          & \multicolumn{1}{c|}{35.6}                                    & \multicolumn{1}{c|}{\textbf{55.8}}                             & \multicolumn{1}{c|}{\textbf{63.4}}                                & \multicolumn{1}{c|}{\textbf{50.9}}                            & \multicolumn{1}{c|}{40.1}                                    & \multicolumn{1}{c|}{\textbf{63.8}}                             & \multicolumn{1}{c|}{\textbf{70.1}}                                & 55.6             \\
\multicolumn{1}{|l|}{\multirow{-2}{*}{\cellcolor[HTML]{EFEFEF}\textbf{\llama}}}  & \cellcolor[HTML]{EFEFEF}\textbf{\sysname}                   & \multicolumn{1}{c|}{50.5}                                & \multicolumn{1}{c|}{\textbf{37.8}}                       & \multicolumn{1}{c|}{\textbf{49.0}}                       & \multicolumn{1}{c|}{44.2}                                & \multicolumn{1}{c|}{58.7}                                & \multicolumn{1}{c|}{\textbf{44.6}}                       & \multicolumn{1}{c|}{\textbf{56.4}}                       & \textbf{53.2} & \multicolumn{1}{c|}{\textbf{44.4}}                           & \multicolumn{1}{c|}{52.9}                                      & \multicolumn{1}{c|}{58.9}                                         & \multicolumn{1}{c|}{48.4}                                     & \multicolumn{1}{c|}{\textbf{55.5}}                           & \multicolumn{1}{c|}{61.0}                                      & \multicolumn{1}{c|}{62.0}                                         & \textbf{56.3}    \\ \hline
\end{tabular}
}
\caption{Execution accuracy on SMCalflow across different code languages and GeoQuery across different types of test splits while increasing M from 5 to 10, and increasing size of training pool $\train$ (top and bottom).  Please see Sec~\ref{sec:data} for reasons for some missing numbers. The number of instances denote size of $\train$ and $\AugD$ separated by '/'. The broad trend is that accuracy gains of \sysname\ is higher for larger $M$ and smaller $\train$.}
\label{tab:smcal_geo_var_m_var_t}
\end{table*}

\paragraph{Evaluation Metric}
We evaluate our approach using execution accuracy (EX), which measures the correctness of the outputs produced by the predicted programs as well as exact match (EM) accuracy depending on the dataset. 
Following ~\cite{bogin-etal-2024-leveraging}, we compute EX by comparing the execution results of predicted and gold programs. For GeoQuery, we compare answers returned by generated programs to those generated by gold programs, while for SMCalFlow, we compare the state (i.e., calendar events) of the environments after execution. 
In case of MTOP, we measure the exact match accuracy.

\paragraph{Baseline}
Our method of augmenting the training set with decompositions of the original question is largely orthogonal to the algorithm used for selecting in-context examples for a test question $\vx$.  Therefore, as a baseline, we choose the state-of-the-art algorithm \coverLS~\cite{LevyBB23} for selecting examples from the original training set.  We  used the official code released by the authors\footnote{\url{https://github.com/itayle/diverse-demonstrations}}. 

\begin{table}[th]
\centering\adjustbox{max width=.8\hsize}{
\begin{tabular}{|l|l|c|c|}
\hline
\textbf{LLM} & \textbf{ICL type} & EN & HI \\ \hline
\multirow{3}{*}{\chatgpt}   & \baseline      & 42.4    & 43.4    \\ 
                            & \indp      & 43.4    & 45.0    \\ 
                            & \sysname & \textbf{45.8 }   & \textbf{50.2 }   \\ \hline
\multirow{3}{*}{\gpt}       & \baseline      & 54.1    & 49.6    \\ 
                            & \indp      & 53.3    & 48.4    \\ 
                            & \sysname & \textbf{54.2}    & \textbf{51.8 }   \\ \hline
\multirow{3}{*}{\mistral}   & \baseline      & 45.8    & 32.7    \\ 
                            & \indp      & 46.8    & 35.8    \\ 
                            & \sysname & 46.6   & \textbf{37.2}    \\ \hline
\multirow{3}{*}{\llama}     & \baseline      & 48.2    & 28.5    \\ 
                            & \indp      & 47.5    & 29.9    \\ 
                            & \sysname & 46.4    & \textbf{34.2}    \\ \hline

\end{tabular}}
\caption{Comparison of Independent Vs Subsumed sub-utterance generation for $M=5$, training split $\train=$ \tfive ( SMCalFlow ) on execution accuracy. Observe how \indp\ provides almost no gains over the baseline \baseline\ in spite of including exactly the same set of sub-trees in $\AugD$. Thus, \sysname's method of generating sub-utterances subsumed by the original utterance is a key reason for its gains. }
\label{tab:decompVariant}
\end{table}

\begin{table*}[htbp]
\centering
\adjustbox{max width=.9\hsize}{%
\setlength{\tabcolsep}{3pt}
\begin{tabular}{|l|l|cccc|cccc|cccc|cccc|}
\hline
\multirow{2}{*}{\textbf{LLM}} & \multirow{2}{*}{\textbf{ICL type}} & \multicolumn{4}{c|}{\textbf{EN}} & \multicolumn{4}{c|}{\textbf{HI}} & \multicolumn{4}{c|}{\textbf{FR}} & \multicolumn{4}{c|}{\textbf{RU}} \\ \cline{3-18}
& & 1 & 2 & 3 & 4 & 1 & 2 & 3 & 4 & 1 & 2 & 3 & 4 & 1 & 2 & 3 & 4 \\
\hline
\multirow{2}{*}{\chatgpt} & \baseline & 50.0 & 52.1 & 36.6 & 53.7 & 50.0 & 48.9 & 40.3 & 44.4 & 50.0 & 45.7 & 37.4 & 31.5 & 50.0 & 54.3 & 45.8 & 44.4 \\
& \sysname & 50.0 & 51.2 & \textbf{42.9} & 42.6 & 50.0 & \textbf{57.5} & \textbf{46.2} & \textbf{46.3} & 50.0 & \textbf{54.4} & \textbf{41.9} & \textbf{33.3} & 50.0 & \textbf{55.3} & \textbf{46.0} & \textbf{55.6} \\
\hline
\multirow{2}{*}{\gpt} & \baseline & 50.0 & 63.9 & 48.9 & 48.1 & 50.0 & 58.5 & 45.0 & 42.6 & 50.0 & 60.4 & 49.0 & 68.5 & 50.0 & 63.0 & 49.6 & 57.4 \\
& \sysname & 50.0 & 62.9 & \textbf{49.8} & 44.4 & 50.0 & \textbf{64.2} & \textbf{45.5} & 37.0 & 50.0 & \textbf{61.0} & \textbf{50.5} & 55.6 & 50.0 & 62.9 & \textbf{52.1} & 53.7 \\
\hline
\multirow{2}{*}{\mistral} & \baseline & 0.0 & 47.5 & 45.1 & 44.4 & 0.0 & 33.9 & 32.7 & 22.2 & 0.0 & 46.3 & 42.3 & 40.7 & 0.0 & 38.4 & 35.5 & 37.0 \\
& \sysname & 0.0 & 44.0 & \textbf{48.4} & 44.4 & \textbf{50.0} & \textbf{39.0} & \textbf{37.3} & 11.1 & \textbf{50.0} & \textbf{50.1} & \textbf{44.0} & 38.9 & 0.0 & \textbf{40.3} & \textbf{40.0} & 31.5 \\
\hline
\multirow{2}{*}{\llama} & \baseline & 0.0 & 47.6 & 48.5 & 53.7 & 0.0 & 32.9 & 27.4 & 1.9 & 0.0 & 44.3 & 43.4 & 40.7 & 50.0 & 43.1 & 37.6 & 29.6 \\
& \sysname & 0.0 & 45.4 & 46.9 & 53.7 & 0.0 & \textbf{39.3} & \textbf{32.4} & \textbf{16.7} & \textbf{50.0} & 44.3 & \textbf{43.7} & 35.2 & 33.3 & \textbf{43.5} & \textbf{38.3} & \textbf{42.6} \\
\hline
\end{tabular}}
\caption{Execution accuracy broken down by depth of the test tree. We consider depth values 1, 2, 3, 4. Almost 30\% of the test trees are of depth 2, and 55\% of depth 3. For a mainstream language like EN, depth 2 trees seem easy enough for the LLM, with most gains from \sysname\ on trees of depth 3.  For an unfamiliar language like HI, depth 2 trees also benefit substantially from \sysname, since interference is likely highest for small trees.}
\label{tab:breakdown}
\end{table*}

\begin{table}[ht]
\centering\adjustbox{max width=.8\hsize}{
\begin{tabular}{|l|c|c|c|}
\hline
\multirow{2}{*}{\textbf{LLM}} & \multicolumn{3}{c|}{\textbf{ICL type}} \\ \cline{2-4}
& \textbf{CoverLS} & \textbf{Mixtral8x22b} & \textbf{GPT4} \\ \hline
{\chatgpt}   & 43.4  & 49.9    & \textbf{50.2}    \\  
{\gpt}       & 49.6  & 51.1    & \textbf{51.8}    \\  
{\mistral}   & 32.7  & 35.1    & \textbf{37.2 }   \\  
{\llama}     & 28.5  & \textbf{34.5 }   & 34.2    \\ \hline
\end{tabular}}
\caption{Comparison of GPT4\_0125 vs Mixtral8x22b decomposition. Even with decompositions from a smaller LLM, \sysname\ improves over baseline.}
\label{tab:utterLLM}
\end{table}

\begin{table}[th]
    \centering\adjustbox{max width=.8\hsize}{
        \begin{tabular}{|l|l|c|c|}
            \hline
            \textbf{LLM}      & \textbf{ICL type}      & EN   & HI   \\ \hline
            \chatgpt              & \sysname              & \textbf{45.8}          & \textbf{50.2}          \\ 
                                & WholeExamples         & 42.5          & 44.7          \\ \hline
            \gpt               & \sysname              & \textbf{54.2}          & \textbf{51.8}          \\ 
                                & WholeExamples         & 50.6          & 48.4          \\ \hline
            \mistral          & \sysname              & \textbf{46.6}          & \textbf{37.2}          \\ 
                                & WholeExamples         & 45.7          & 34.5          \\ \hline
            \llama           & \sysname              & 46.4          & \textbf{34.2}          \\ 
                                & WholeExamples         & \textbf{46.8}          & 27.3          \\ \hline
        \end{tabular}
    }
    \caption{Comparison of \sysname\ and WholeExamples on SMCalFlow (training split: $\train=$ \tfive, $M=5$).}
    \label{tab:vsWholeExs}
\end{table}

\section{Results}

We first present the performance of the proposed algorithm relative to various baselines and then discuss various ablations to dissect the reason for the gains.

\paragraph{SCUD4ICL vs.  Baselines}
In Tables~\ref{tab:overall}, \ref{tab:mtop}, and \ref{tab:sm_otherbaselines}, we compare accuracy of \sysname\ with the baseline \baseline\  across various datasets  and splits discussed earlier. Results with additional splits are presented in the Appendix \ref{appendix:add_splits}. 
Based on the results, we make the following observations. \begin{enumerate}[(1),wide,font=\bfseries,labelindent=0pt]
\item First, across both datasets and varying test-train splits, \sysname\ provides much higher accuracy than \baseline\ and significantly outperforms  DPP and TOPK  \citep{LevyBB23}. The main reason for the gains over \baseline\ is the augmentation of the training pool with decomposed training instances since the algorithm used for selecting the $M$ instances are \todo{@SS debatable} largely similar.
\item As expected, just by changing the language in the names of clauses, accuracy varies across all LLMs.  For a low resource language such as Hindi (Hi), \sysname\ provides much higher gains than on  English.  For example, with \chatgpt\ and \llama\ we observe more than 6\% absolute jump in accuracy on SMCalFlow-Hi.

\item \sysname\ provides better generalization when train-test splits differ in length as seen from Geoquery's \textit{length} test split. Even for \gpt\ we observe a jump in accuracy from 73.6 to 81, and for \mistral\ accuracy jumps from 32.3\% to 50.6\%. 
\item \sysname\ also generalizes better when there is discrepancy in the template of the test and train trees as seen in  Geoquery's \textit{tmcd} test split.  For \chatgpt\ accuracy jumps from 70.9\% to 82\%, and for \mistral\ the jump is from 49.4\% to 73.9\%.
\end{enumerate}



\paragraph{Decomposition variants}
We analyze whether the gains are just because of augmenting the train pool $\train$ with decomposed sub-trees, or whether the quality of the sub-utterance was the key reason.
In Table~\ref{tab:decompVariant} we demonstrate the impact of our method of generating subsumed sub-utterances by comparing with independently generated sub-utterances as described in Section~\ref{sec:Decomp}.   We observe that with independent sub-utterances, the performance is similar to the original un-augmented data.  Only with our subsumed sub-utterance, do we get the accuracy gains seen above.  The examples in Figure~\ref{fig:lang_pres_ablation} illustrate that independent utterances often do not appear natural enough, and they possibly fail to match test utterances. Another question is whether sub-utterances generated by a lower capacity open LLM such as Mixtral:8x22b~\citep{mistral2024mixtral8x22b} 
compare with those from GPT4\_0125. In Table~\ref{tab:utterLLM} we show accuracy with SMCalFlow-Hi.  Observe that even with Mixtral:8x22b-generated sub-utterances, \sysname\ provides adequate gains over the baseline, although the gains are greater with \gpt.

\paragraph{Impact of Fragment Selection}
To examine the impact of fragment selection in in-context examples, \sysname-selected fragments were compared with the corresponding full ICE examples, referred to as WholeExamples. Both approaches were evaluated using the same set of examples to ensure a controlled comparison.

The results shown in Table~\ref{tab:vsWholeExs} indicate that \sysname\ consistently achieves superior or comparable performance compared to WholeExamples across all evaluated models and languages. This demonstrates that the removal of irrelevant parts in ICEs reduces distractions and contributes to improved predictive performance.

\paragraph{Accuracy gains for different test tree sizes}
We show a breakdown of the accuracy gains across depth of the test trees in Table~\ref{tab:breakdown}.  For a mainstream language like English depth 2 trees may be easy enough, and we do not see much gains with \sysname.  For low resource language like Hindi,
biggest gains are obtained from shorter trees (depth=2) that are more subject to interference from irrelevant clauses in the decomposition.  
\begin{figure}
    \centering
    \includegraphics[width=0.8\linewidth]{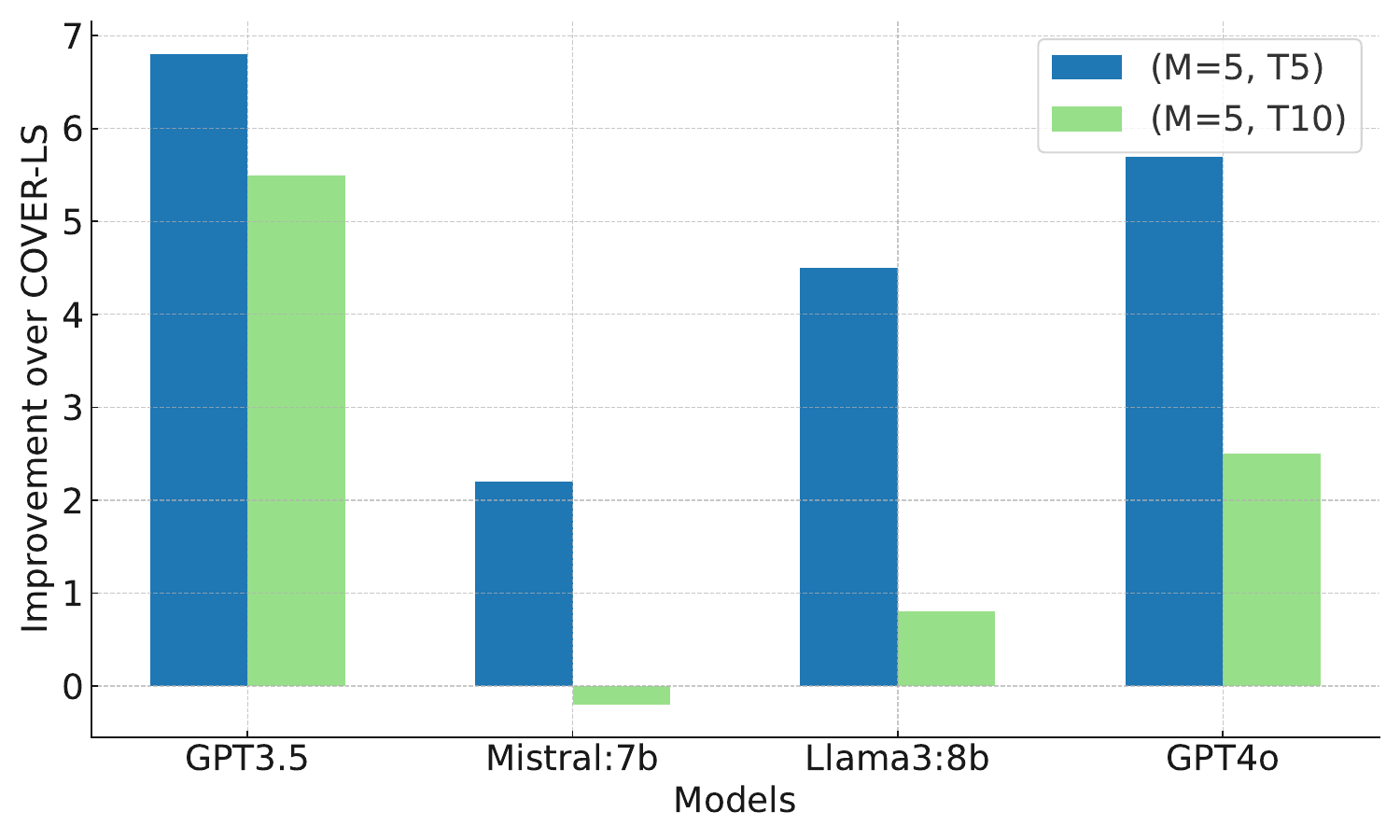}
    \caption{Accuracy gains of \sysname\ over baseline for SMCalFlow-Hi version for two different training pool sizes pointing to higher gains for a smaller pool.}
    \label{fig:barTS}
\end{figure}

\paragraph{Increasing $M$ and training pool $\train$}
In Table~\ref{tab:smcal_geo_var_m_var_t} we show accuracy with $M$, the size of the in-context set, increasing from 5 to 10, and two different training sizes. 
A rough trend to observe is that for larger $M$, \sysname's gains over baseline increases.  Consider, for instance Geoquery's \textit{iid} split on the \geoA\ train set. 
Across all LLMs, \sysname\ provides higher gains with $M=10$ than with $M=5$ on this train-test split.  Such a trend is explained by the fact that baseline is more likely to include irrelevant clauses for large $M$, and  our decomposition is able to eliminate them.

When the training pool $\train$ size increases with a fixed $M$, the baseline is likely to find increasingly relevant instances for ICL.  Thus, we expect the impact of decomposition to be higher when the training pool is smaller.  In Table~\ref{tab:smcal_geo_var_m_var_t} we observe that relative gains are higher in the top-half compared to the bottom half.  We zoom in further on SMCalFlow-Hi for $M=5$ in Figure~\ref{fig:barTS}.  Observe the increased relative gains $\train=\tfive$ with 71 instances compared to $\train=\tten$ with 116 instances.

In summary, our experiments show that while \sysname\ provides overall gains over baseline, it is particularly useful when the code language is less familiar, there is mismatch in the train and test distribution in terms of code length and template, IC budget is large, and the training pool is small.




\section{Conclusion}
\label{sec:End}
We introduced \sysname, a semantic interpreter that incorporates a new paradigm of fragmenting structured programs 
from the pool of available ICEs, and instructing an LLM to translate these program fragments back to natural language utterances.  This forms a larger ICE pool including decomposed (utterance, program) pairs.  Test instance-guided diverse ICE selection from this enlarged pool improves semantic interpretation accuracy for a number of benchmarks.


\section{Limitations}
\label{sec:Limit}

Although use of commercial LLMs as network services is widespread in this nature of research, it severely reduces reproducibility.
As one example, we do not understand how the LLM implements utterance subsumption.  Should the LLM change significantly owing to instruction tuning, our results may change drastically.
Further experiments with prompt/prefix tuning, and/or setting up a smaller in-house LLM and adapting it to our task, would be of future interest.

\bibliography{ICL, FoCL}

\clearpage
\appendix


\begin{center}
\Large\bfseries\ztitle \\ (Appendix)
\end{center}

\section{Additional Experimental Results}

\subsection{Exact Match Accuracy}

\begin{table}[ht]
    \centering\adjustbox{max width=.9\hsize}{
    \begin{tabular}{|l|l|c|c|c|c|}
        \hline
        \textbf{LLM}       & \textbf{ICL Type}   & EN   & HI   & FR   & RU   \\ \hline
        \chatgpt    & \baseline     & 18.0          & 17.0          & 14.4          & 17.6          \\ 
                    & \sysname      & \textbf{21.1} & \textbf{19.5} & \textbf{18.3} & \textbf{19.6}          \\ \hline
        \gpt        & \baseline     & \textbf{18.0} & 12.3          & \textbf{14.9} & 16.8          \\ 
                    & \sysname      & 17.5          & \textbf{12.9} & 14.2          & \textbf{16.8} \\ \hline
        \mistral    & \baseline     & \textbf{13.8} & \textbf{7.7}  & \textbf{11.7} & 9.5           \\ 
                    & \sysname      & 13.3          & 6.9           & 11.3          & \textbf{9.7}  \\ \hline
        \llama      & \baseline     & \textbf{13.0} & 5.8           & \textbf{10.3} & 8.5           \\ 
                    & \sysname      & 11.6          & \textbf{7.7}  & 10.1          & \textbf{10.7} \\ \hline
    \end{tabular}}
    \caption{Exact Match accuracy for Table~\ref{tab:overall}, training split $\train=$ \tfive\ (SMCalFlow)}
    \label{appendix:sm_em}
\end{table}

Table~\ref{appendix:sm_em} compares \sysname\ and CoverLS in terms of exact match accuracy. While \sysname\ consistently outperformed \baseline\ in execution accuracy (Table~\ref{tab:overall} in the paper), its performance in terms of exact match accuracy is mixed. \sysname\ excels on older models like \chatgpt\ and in low-resource languages (e.g., Hindi and Russian), but the lower scores in other cases can be attributed to its approach of assembling code using information in sub-fragment examples. This strategy often produces semantically correct outputs that differ syntactically from the gold standard, leading to lower exact match scores despite maintaining functional correctness. 

For example, the query "Hi, can you reschedule me a meeting with Ruth on Monday?" can be translated as either "CreateEvent( AND( starts\_at( NextDOW( MONDAY ) ), with\_attendee( Ruth ) ) )" or "CreateEvent( AND( with\_attendee( Ruth ), starts\_at( NextDOW( MONDAY ) ) ) )". If the former is the gold annotation, the latter is still correct in terms of execution accuracy but fails on exact match.

\subsection{Variability in Execution Accuracy Results}

\begin{table}[ht]
    \centering\adjustbox{max width=.9\hsize}{
    \begin{tabular}{|l|l|c|c|c|c|}
        \hline
        \textbf{LLM}      & \textbf{ICL type} & EN   & HI   & FR   & RU   \\ \hline
        \chatgpt              & \baseline           & 0.15          & 0.25          & 0.75          & 1.0           \\ 
                            & \sysname          & 0.30          & 0.29          & 0.59          & 0.46          \\ \hline
        \gpt               & \baseline           & 2.24          & 1.42          & 0.38          & 0.76          \\ 
                            & \sysname          & 0.46          & 0.25          & 0.17          & 0.84          \\ \hline
        \mistral          & \baseline           & 1.08          & 1.10          & 1.16          & 0.75          \\ 
                            & \sysname          & 0.61          & 0.55          & 0.68          & 0.47          \\ \hline
        \llama           & \baseline           & 0.29          & 0.86          & 0.93          & 0.31          \\ 
                            & \sysname          & 0.96          & 0.17          & 1.01          & 0.83          \\ \hline
    \end{tabular}}
    \caption{Standard deviation for Table~\ref{tab:overall}, training split $\train=$ \tfive\ (SMCalFlow)}
    \label{appendix:std}
\end{table}

The reported results account for randomness, with all values representing the average of three runs. Standard deviations for Table~\ref{tab:overall} are provided in Table~\ref{appendix:std}.

\subsection{Evaluation with Additional Data Splits}
Table~\ref{appendix:add_splits} presents the results comparing \baseline\ and \sysname\ across the random training subsplits of SMCalFlow (as well as the complete training split for GeoQuery), evaluated using different LLMs and languages. These results indicate that \sysname\ consistently outperforms or matches \baseline\, leveraging its ability to identify highly relevant sub-fragments, even while using fewer tokens. Note: Random split 1 contains 100 instances and 825 decompositions, whereas random split 2 contains 454 instances and 3,758 decompositions.

\begin{table*}
    \centering
    \adjustbox{max width=.8\hsize}{%
    \begin{tabular}{|l|l|c|c|c|c|c|c|c|c|}
        \hline
        \multirow{3}{*}{\textbf{LLM}} & \multirow{3}{*}{\textbf{ICL type}} & \multicolumn{4}{c|}{\textbf{SMCalFlow}} & \multicolumn{4}{c|}{\textbf{GeoQuery}} \\ \cline{3-10} 
        &  & \multicolumn{2}{c|}{Split 1} & \multicolumn{2}{c|}{Split 2} & Length & i.i.d & Template & TMCD \\ \cline{3-10} 
                                       &                                  & EN & HI & EN & HI & \multicolumn{4}{c|}{EN} \\ \hline
        \chatgpt                      & \baseline                        & 43.7       & 48.8       & 48.3       & 56.4       & 66.4   & 86.1  & 80.7     & 73.0 \\ 
                                       & \sysname                         & \textbf{47.4} & \textbf{50.8} & \textbf{53.0} & \textbf{57.7} & \textbf{77.0} & \textbf{87.1} & \textbf{86.7} & \textbf{80.0} \\ \hline
        \mistral                      & \baseline                        & 49.1       & 36.2       & \textbf{55.4} & 39.2       & 46.1   & 71.4  & 64.8     & 61.2 \\ 
                                       & \sysname                         & \textbf{51.2} & \textbf{36.4} & 53.3       & \textbf{42.3} & \textbf{64.8} & \textbf{74.6} & \textbf{70.8} & \textbf{65.5} \\ \hline
        \llama                        & \baseline                        & 50.6       & 30.5       & \textbf{55.4} & 36.7       & 45.2   & \textbf{63.9} & 64.8     & 56.7 \\ 
                                       & \sysname                         & \textbf{51.5} & \textbf{34.0} & 55.3       & \textbf{38.4} & \textbf{58.8} & 62.5  & \textbf{71.1} & \textbf{58.5} \\ \hline
    \end{tabular}}
    \caption{Execution accuracy on random training splits of SMCalFlow and full training split of GeoQuery ($M=5$).}
    \label{appendix:add_splits}
\end{table*}

\subsection{Evaluation of Decomposed Examples in SMCalFlow}

An evaluation of the generated sub-utterances was conducted as part of the manual assessment of the SMCalFlow $\train=$ \tfive\ training split. The results are as follows:

\begin{itemize}
    \item Original questions: 71
    \item Total decompositions generated: 296
    \item Incorrectly generated decompositions: 30
\end{itemize}

This corresponds to a decomposition error rate of 10\% of the overall decompositions, with some errors being relatively minor. Despite these errors, experiments demonstrate that the use of decomposed examples leads to improved overall accuracy.

\section{SCUD4ICL - Optimization Perspective}
\label{appendix:optimization}
\sysname\ selects examples for in-context learning by solving an optimization problem that balances relevance the test query $\vx_0$ and diversity within the selected set $S$, constrained by a budget $M$. Our ICE selection follows a greedy approach but can be viewed as an instantiation of the below  optimization problem:


\begin{align*}
&\underset{S: |S| \leq M}{\max} \Biggl[ \sum_{(n,k)\in S} \Bigl( \alpha \cdot R(\vx_{n,k}, \vx_0) \\
&\qquad\qquad + \beta \cdot D(\vx_{n,k}, S) \Bigr)\Biggr]
\end{align*}

Here,  relevance $R(\vx_{n,k}, \vx_0)$ measures the similarity between $\vx_{n,k}$ and $\vx_0$ and diversity $D(\vx_{n,k}, S)$ is defined as the \textit{negative minimal similarity} with other members of $S$:
\begin{equation*}
D(\vx_{n,k}, S) = -\min_{(m,l) \in S \setminus \{ (n,k) \} }\text{sim}(\vx_{n,k}, \vx_{m,l}),
\end{equation*}
where $\text{sim}(\vx_{n,k}, \vx_{m,l})$ measures structural and semantic similarity.\\

\noindent \textbf{Specific Design Choices in \sysname}\\
\begin{itemize}
    \item \textit{Relevance ($R$)} is computed using BM25 to assess semantic alignment with $\vx_0$.
    \item \textit{Diversity ($D$)} penalizes similarity to examples already selected, disqualifying candidates with overlaps of their anonymized versions or ancestor-descendant relationships. 
    \item \textit{Token-based Filtering} enables selection of  candidates $x_{n,k}$ that contain tokens from $x_0$ not yet covered, improving efficiency.
\end{itemize}

\noindent To summarize, \sysname~ integrates ideas from Diverse Demonstrations \cite{LevyBB23} and DPP-based selection \cite{Ye2023CEIL}, while applying structural constraints (e.g., template matching and hierarchical relationships) to improve selection from a decomposed example pool.

\section{Prompts}
In Figures~\ref{prompt:semparse:en} and \ref{prompt:semparse:hi} we sketch the prompt we used to obtain semantic interpretation in the SMCalFlow dataset on two code languages En and Hi respectively.  

\begin{figure*}[h]
    \begin{center}
        \begin{tcolorbox}[width=0.9\hsize, colframe=black, colback=gray!5, boxrule=0.5pt, left=0.2in, right=0.2in, top=0.2in, bottom=0.2in, label=qd_prompt_two, sharp corners=southwest]
        \parskip 1.5ex \ttfamily \small
        Given the following data structures and functions:
        
        \begin{verbatim}
        FindTeamOf  # given a person name or ID, ...
        has_subject  # given a string, returns an ...
        starts_at    # given a datetime clause, ...
        CreateEvent  # given multiple event clauses ... 
        ...
        ...
        \end{verbatim}
        
        Your task is to write DSL code for the given question.
        
        \textbf{Note:}\\
        1. Do not use any external libraries/functions.\\
        2. Strictly adhere to the provided operators.
        \end{tcolorbox}
        \caption{Instruction to LLM for EN code generation (SMCalFlow).}
        \label{prompt:semparse:en}
    \end{center}
\end{figure*}

\begin{figure*}[ht]
    \begin{center}
        \begin{tcolorbox}[width=0.9\hsize, colframe=black, colback=gray!5, boxrule=0.5pt, left=0.2in, right=0.2in, top=0.2in, bottom=0.2in, label=qd_prompt_three, sharp corners=southwest]
        \parskip 1.5ex \ttfamily \small
        Given the following data structures and functions:
        
        \begin{verbatim}
        DalKhojen  # given a person name or ID, ...
        VishayHai  # given a string, returns an ... 
        SePrarambh  # given a datetime clause, ... 
        KaryakramBanao  # given multiple event ...
        ...
        ...
        \end{verbatim}
        
        Your task is to write DSL code for the given question.
        
        \textbf{Note:}\\
        1. Do not use any external libraries/functions.\\
        2. Strictly adhere to the provided operators.
        \end{tcolorbox}
        \caption{Instruction to LLM for HI code generation (SMCalFlow).}
        \label{prompt:semparse:hi}
    \end{center}
\end{figure*}

\begin{figure*}[t]
        \begin{tcolorbox}[
            width=\hsize,
    	colframe=black,
    	colback=gray!5,
    	boxrule=0.5pt,
    	left=0.2in,
    	right=0.2in,
    	top=0.2in,
    	bottom=0.01in,
    	label=qd_prompt_four,
    	sharp corners=southwest]
            \ttfamily\tiny 
            \begin{lstlisting}[breaklines=true]
                
                FindTeamOf  # given a person name or id, returns a pseudo-person representing the team of that person
                FindReports  # given a person name or id, returns a pseudo-person representing the reports of that person
                FindManager  # given a person name or id, returns the manager of that person
                
                with_attendee  # given a person name or id, returns a clause to match or create an event with that person as an attendee
                avoid_attendee  # given a person name or id, returns an event clause to avoid that attendee when creating an event
                has_subject  # given a string, returns an event to match or create an event with that subject
                at_location  # given a string, returns an event clause to match or create an event at that location
                starts_at  # given a datetime clause, returns an event clause to match or create an event starting at that time
                ends_at  # given a datetime clause, returns an event clause to match or create an event ending at that time
                has_duration  # given a time unit value, returns an event clause to match or create an event with that duration
                has_status  # given a ShowAsStatus value, returns an event clause to match or create an event with that status
                
                # the following operators return datetime clauses and accept no arguments
                Afternoon
                Breakfast
                Brunch
                Dinner
                Early
                EndOfWorkDay
                Evening
                FullMonthofMonth
                FullYearofYear
                LastWeekNew
                Late
                LateAfternoon
                LateMorning
                Lunch
                Morning
                NextMonth
                NextWeekend
                NextWeekList
                NextYear
                Night
                Noon
                Now
                SeasonFall
                SeasonSpring
                SeasonSummer
                SeasonWinter
                ThisWeek
                ThisWeekend
                Today
                Tomorrow
                Yesterday
                
                # general date time clauses
                DateTime  # given either a datetime clause representing a date and/or a time operator representing a time, returns a datetime clause
                Date  # given a date or dayofweek, returns a date
                DayOfWeek  # given a day of week string, returns a time clause
                NextDOW  # given a day of week string, returns a time clause for the next occurrence of that day of week
                MD  # given a month and day as arguments, returns a date clause
                MDY  # given a month, day, and year as arguments, returns a date clause
                
                # given a value, the following operators return datetime clauses according to the given value
                toMonth
                toFourDigitYear
                HourMinuteAm
                HourMinutePm
                NumberAM
                NumberPM
                
                # given a datetime clause, the following operators modify the clause and return a datetime clause according to the modification
                OnDateAfterTime
                OnDateBeforeTime
                AroundDateTime
                
                # given either a number or the operators Acouple/Afew, all the following operators return time unit values according to the given unit
                toDays
                toHours
                toMinutes
                
                # these operators can be used to create time unit values instead of using integer values
                Acouple
                Afew
                
                ShowAsStatus  # enumeration of possible event statuses (Busy, OutOfOffice)
                
                AND  # combines multiple event clauses together
            \end{lstlisting}
        \end{tcolorbox}
    \caption{Exhaustive list of operators for EN code generation (SMCalFlow).}
    \label{schema:en}
\end{figure*}


\begin{figure*}[t]
    \begin{tcolorbox}[
        width=\hsize,
	colframe=black,
	colback=gray!5,
	boxrule=0.5pt,
	left=0.2in,
	right=0.2in,
	top=0.2in,
	bottom=0.01in,
	label=qd_prompt_five,
	sharp corners=southwest]
        \ttfamily\tiny 
        \begin{lstlisting}[breaklines=true]
            
            DalKhojen  # given a person name or id, returns a pseudo-person representing the team of that person
            ReportDhoondho  # given a person name or id, returns a pseudo-person representing the reports of that person
            PrabandhakKhojen  # given a person name or id, returns the manager of that person
            
            InSahbhagiyonKeSaath  # given a person name or id, returns a clause to match or create an event with that person as an attendee
            InSahbhagiyonKeBina  # given a person name or id, returns an event clause to avoid that attendee when creating an event
            VishayHai  # given a string, returns an event to match or create an event with that subject
            IsSthanPar  # given a string, returns an event clause to match or create an event at that location
            SePrarambh  # given a datetime clause, returns an event clause to match or create an event starting at that time
            PeSamapt  # given a datetime clause, returns an event clause to match or create an event ending at that time
            AvdhiHai  # given a time unit value, returns an event clause to match or create an event with that duration
            SthitiHai  # given a ShowAsStatus value, returns an event clause to match or create an event with that status
            
            # the following operators return datetime clauses and accept no arguments
            DopaharBaad
            Naashta
            DerNashta
            RaatKaBhojan
            Jaldi
            KaryaDivasSamapt
            Shaam
            MaahKaPuraMaah
            VarshKaPurnaVarsh
            PichleHafteNaya
            Der
            DerDopahar
            DerSubah
            DopaharKaBhojan
            Subah
            AglaMaah
            AglaSaptahant
            AgleHafteKiSuchi
            AglaVarsh
            Raat
            Dopahar
            Abhi
            Patjhad
            Vasant
            Grishm
            Shishir
            IsHafte
            IsSaptahant
            Aaj
            Kal
            BitaKal
            
            # general date time clauses
            DinankSamayVarg  # given either a datetime clause representing a date and/or a time operator representing a time, returns a datetime clause
            DinankVarg  # given a date or dayofweek, returns a date
            SaptahKaVarshikDin  # given a day of week string, returns a time clause
            AglaKaryaDiwas  # given a day of week string, returns a time clause for the next occurrence of that day of week
            MahinaDin  # given a month and day as arguments, returns a date clause
            MahinaDinVarsh  # given a month, day, and year as arguments, returns a date clause
            
            # given a value, the following operators return datetime clauses according to the given value
            MaahMein
            ChaarAnkVarsh
            GhantaMinatPoorvahn
            GhantaMinatAparanh
            SankhyaPoorvahn
            SankhyaAparanh
            
            # given a datetime clause, the following operators modify the clause and return a datetime clause according to the modification
            DinankKeBaadSamay
            DinankParSamaySePhele
            SamayDinankKePaas
            
            # given either a number or the operators EkDo/Kuch, all the following operators return time unit values according to the given unit
            DinoMein
            GhantoMein
            MinatoMein
            
            # these operators can be used to create time unit values instead of using integer values
            EkDo
            Kuch
            
            SthitiDikhayein  # enumeration of possible event statuses (Busy, OutOfOffice)
            
            Aur  # combines multiple event clauses together
        \end{lstlisting}
    \end{tcolorbox}
    \caption{Exhaustive list of operators for HI code generation (SMCalFlow).}
    \label{schema:hi}
\end{figure*}


\lstset{
    inputencoding=utf8/latin1,
    extendedchars=true,
    literate={é}{{\'e}}1 {É}{{\'E}}1 {À}{{\`A}}1
}
\begin{figure*}[t]
    \begin{tcolorbox}[
        width=\hsize,
	colframe=black,
	colback=gray!5,
	boxrule=0.5pt,
	left=0.2in,
	right=0.2in,
	top=0.2in,
	bottom=0.01in,
	label=qd_prompt_six,
	sharp corners=southwest]
        \ttfamily\tiny 
        \begin{lstlisting}[breaklines=true]
            
            TrouverÉquipeDe  # given a person name or id, returns a pseudo-person representing the team of that person
            TrouverRapports  # given a person name or id, returns a pseudo-person representing the reports of that person
            TrouverGestionnaire  # given a person name or id, returns the manager of that person
            
            avec_participant  # given a person name or id, returns a clause to match or create an event with that person as an attendee
            éviter_participant  # given a person name or id, returns an event clause to avoid that attendee when creating an event
            a_sujet  # given a string, returns an event to match or create an event with that subject
            à_emplacement  # given a string, returns an event clause to match or create an event at that location
            commence_à  # given a datetime clause, returns an event clause to match or create an event starting at that time
            se_termine_à  # given a datetime clause, returns an event clause to match or create an event ending at that time
            a_durée  # given a time unit value, returns an event clause to match or create an event with that duration
            a_statut  # given a ShowAsStatus value, returns an event clause to match or create an event with that status
            
            # the following operators return datetime clauses and accept no arguments
            AprèsMidi
            PetitDéjeuner
            Brunch
            Dîner
            Tôt
            FinDeJournéeDeTravail
            Soirée
            MoisEntierDuMois
            AnnéeComplèteDeL'Année
            NouvelleDernièreSemaine
            Tard
            FinD'AprèsMidi
            FinDeMatinée
            Déjeuner
            Matin
            MoisProchain
            WeekEndProchain
            ListeProchaineSemaine
            AnnéeProchaine
            Nuit
            Midi
            Maintenant
            Automne
            Printemps
            Été
            Hiver
            CetteSemaine
            CeWeekEnd
            Aujourd'hui
            Demain
            Hier
            
            # general date time clauses
            ClasseDateHeure  # given either a datetime clause representing a date and/or a time operator representing a time, returns a datetime clause
            ClasseDate  # given a date or dayofweek, returns a date
            ClasseJourDeSemaine  # given a day of week string, returns a time clause
            ProchainJourOuvré  # given a day of week string, returns a time clause for the next occurrence of that day of week
            MoisJour  # given a month and day as arguments, returns a date clause
            MoisJourAnnée  # given a month, day, and year as arguments, returns a date clause
            
            # given a value, the following operators return datetime clauses according to the given value
            versMois
            enAnnéeÀQuatreChiffres
            HeureMinuteAM
            HeureMinutePM
            NombreAM
            NombrePM
            
            # given a datetime clause, the following operators modify the clause and return a datetime clause according to the modification
            ÀDateAprèsHeure
            ÀDateAvantHeure
            AutourDateHeure
            
            # given either a number or the operators EkDo/Kuch, all the following operators return time unit values according to the given unit
            enJours
            enHeures
            enMinutes
            
            # these operators can be used to create time unit values instead of using integer values
            UnCouple
            Quelques
            
            AfficherCommeStatut  # enumeration of possible event statuses (Busy, OutOfOffice)
            
            ET  # combines multiple event clauses together
        \end{lstlisting}
    \end{tcolorbox}
    \caption{Exhaustive list of operators for FR code generation (SMCalFlow).}
    \label{schema:fr}
\end{figure*}


\begin{figure*}[t]
    \centering
    \fbox{\includegraphics[width=0.74\paperwidth,height=0.7\paperheight]{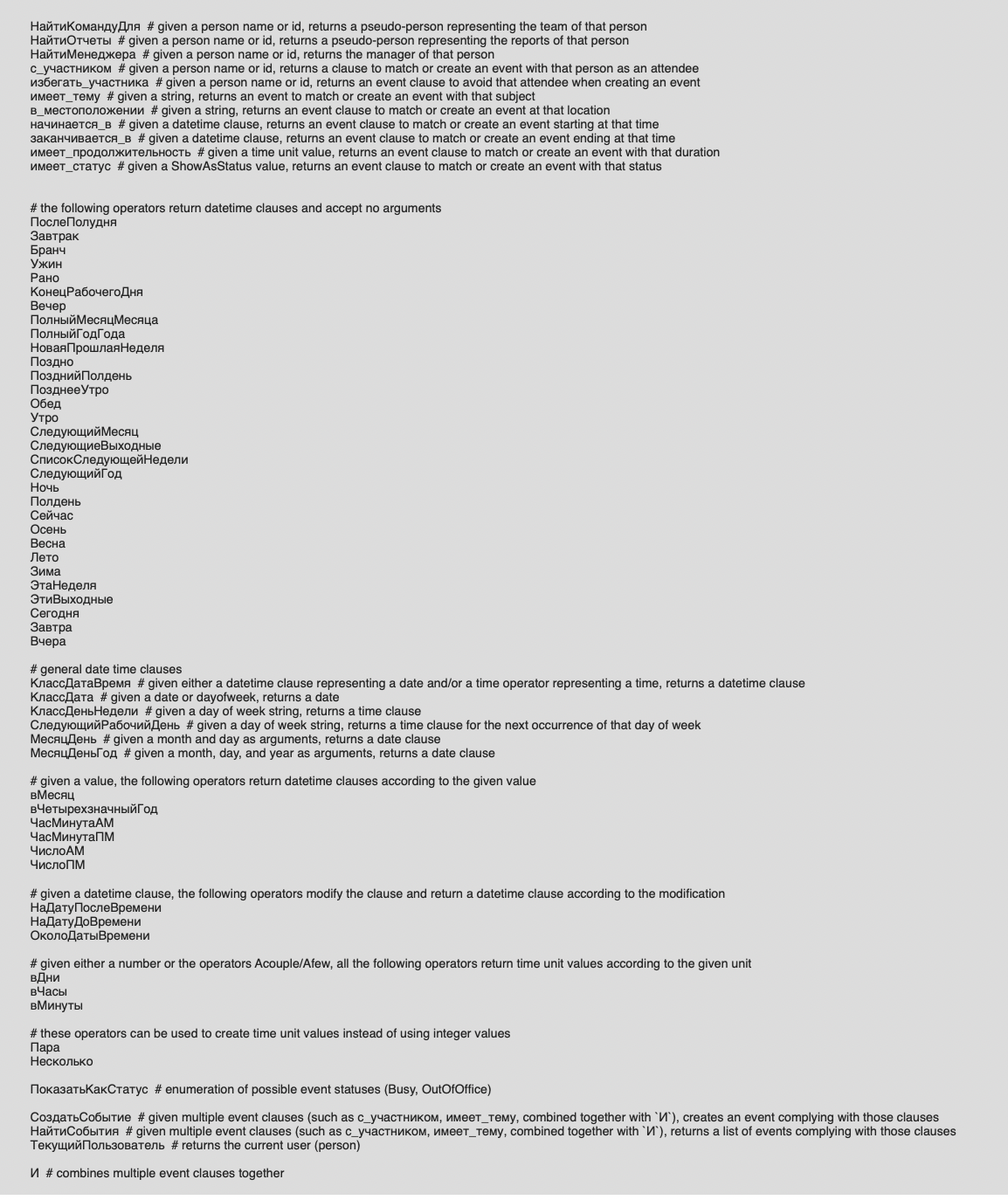}}  
    \label{schema:ru}
\caption{Exhaustive list of operators for RU code generation (SMCalFlow).}
\end{figure*}

\newpage

\lstset{
  basicstyle=\ttfamily\small,
  breaklines=true,
  showstringspaces=false,
  columns=flexible,
  keepspaces=true,
  frame=single,
  xleftmargin=2pt,
  xrightmargin=2pt
}

\tcbset{
  enhanced jigsaw,
  breakable,
  colback=white,
  boxrule=0.5pt,
  sharp corners,
  arc=0mm,
  left=2pt,
  right=2pt,
  top=2pt,
  bottom=2pt
}

\begin{figure*}[t]
    \centering
    \begin{tabular}{|p{\textwidth}|}
        \hline
        \textbf{USER} \\
        \begin{lstlisting}[numbers=none, backgroundcolor=\color{gray!5}]
original question: Add engagement party to monday 9 pm with Abby kim and desi

code: CreateEvent( AND( with_attendee( Abby ) , with_attendee( Kim ) , 
    with_attendee( Desi ) , has_subject( engagement party ) , 
    starts_at( NextDOW( MONDAY ) ) , starts_at( NumberPM( 9 ) ) ) )

decomposition:
        \end{lstlisting} \\
        \hline
        \textbf{ASSISTANT} \\
        \begin{lstlisting}[language=json,numbers=none, backgroundcolor=\color{gray!5}]
{
  "Add engagement party to monday 9 pm with Abby kim and desi": {
    "code": "CreateEvent( AND( with_attendee( Abby ), with_attendee( Kim ),
      with_attendee( Desi ), has_subject( engagement party ),
      starts_at( NextDOW( MONDAY ) ), starts_at( NumberPM( 9 ) ) ) )",
    "decomposition": {
      "Add engagement party with Abby kim and desi": {
        "code": "CreateEvent( AND( with_attendee( Abby ),
          with_attendee( Kim ), with_attendee( Desi ),
          has_subject( engagement party ) ) )",
        "decomposition": {
          "Add event engagement party": {
            "code": "CreateEvent( has_subject( engagement party ) )",
            "decomposition": {}
          },
          "Add event with Abby kim and desi": {
            "code": "CreateEvent( AND( with_attendee( Abby ),
              with_attendee( Kim ), with_attendee( Desi ) ) )",
            "decomposition": {}
          }
        }
      },
      "Add event to monday 9 pm": {
        "code": "CreateEvent( AND( starts_at( NextDOW( MONDAY ) ),
          starts_at( NumberPM( 9 ) ) ) )",
        "decomposition": {}
      }
    }
  }
}
        \end{lstlisting} \\
        \hline
    \end{tabular}
    \caption{An in-context example for query decomposition.}
    \label{fig:qd_ICL_1}
\end{figure*}

\newpage

\end{document}